\pdfoutput=1

\documentclass[11pt]{article}

\usepackage[]{acl}
\usepackage{times}
\usepackage{latexsym}
\usepackage{graphicx}
\usepackage[most]{tcolorbox}
\usepackage{booktabs}
\usepackage{array,multirow}
\usepackage{colortbl}
\usepackage{xcolor}
\usepackage[T1]{fontenc}

\usepackage[utf8]{inputenc}

\usepackage{microtype}

\usepackage{inconsolata}

\title{Towards Smarter Hiring: Are Zero-Shot and Few-Shot Pre-trained LLMs Ready for HR Spoken Interview Transcript Analysis?}

 \author{
    Subhankar Maity$^{1}$ \quad Aniket Deroy$^{1}$ \quad Sudeshna Sarkar$^{1}$ \\
    $^{1}$Indian Institute of Technology Kharagpur (IIT-Kgp) \\
    \texttt{\{subhankar.ai, roydanik18\}@kgpian.iitkgp.ac.in} \\
    \texttt{sudeshna@cse.iitkgp.ac.in}
}

\begin{document}
\maketitle
\begin{abstract}
This research paper presents a comprehensive analysis of the performance of prominent pre-trained large language models (LLMs), including GPT-4 Turbo, GPT-3.5 Turbo, \texttt{text}-\texttt{davinci}-\texttt{003}, \texttt{text}-\texttt{babbage}-\texttt{001}, \texttt{text}-\texttt{curie}-\texttt{001}, \texttt{text}-\texttt{ada}-\texttt{001}, \texttt{llama}-\texttt{2}-\texttt{7b}-\texttt{chat}, \texttt{llama}-\texttt{2}-\texttt{13b}-\texttt{chat}, and \texttt{llama}-\texttt{2}-\texttt{70b}-\texttt{chat}, in comparison to expert human evaluators in \textit{providing scores}, \textit{identifying errors}, and \textit{offering feedback and improvement suggestions} to candidates during mock HR (Human Resources) interviews. We introduce a dataset called \textbf{HURIT} (\underline{\textbf{Hu}}man \underline{\textbf{R}}esource \underline{\textbf{I}}nterview \underline{\textbf{T}}ranscripts), which comprises 3,890 HR interview transcripts sourced from real-world HR interview scenarios. Our findings reveal that pre-trained LLMs, particularly GPT-4 Turbo and GPT-3.5 Turbo, exhibit commendable performance and are capable of producing evaluations comparable to those of expert human evaluators. Although these LLMs demonstrate proficiency in \textit{providing scores} comparable to human experts in terms of human evaluation metrics, they frequently fail to \textit{identify errors} and \textit{offer specific actionable advice} for candidate performance improvement in HR interviews. Our research suggests that the current state-of-the-art pre-trained LLMs are not fully conducive for automatic deployment in an HR interview assessment. Instead, our findings advocate for a human-in-the-loop approach, to incorporate manual checks for inconsistencies and provisions for improving feedback quality as a more suitable strategy.



\end{abstract}

\section{Introduction}

Education and communication skills are critical in the context of a fresher's job interview, especially during the HR round \citep{r6}. The interconnectedness of education and communication skills significantly impacts the interview process \citep{r5}. Education goes beyond subject-specific knowledge; it helps develop soft skills, including communication \citep{r7}. In the HR round, the emphasis is on how the educational experience of a candidate has improved their ability to communicate effectively, both verbally and in writing \citep{r6}. 


For L2\footnote{\url{https://tinyurl.com/255p2355}} English speakers, the need to assess soft skills, especially English communication proficiency, is even more pronounced than for L1 English speakers \citep{r8}. The challenges and nuances of expressing oneself in a non-native language add a layer of complexity to the evaluation of communication proficiency, specifically in the English language \citep{r9}. In contrast, L1 English speakers navigate the job interview landscape with the advantage of a native language foundation. According to \citet{r10}, a poor command of English ranks among the top five reasons why fresh graduates are not hired.

This study delves into the evaluation of candidate responses to HR questions in the context of a job interview scenario. The assessment focuses primarily on communication skills, with a specific focus on English proficiency for L2 English speakers. Traditionally, evaluating communication skills through human expert evaluation is expensive and time-consuming, particularly for evaluating freshers seeking job opportunities.


The emergence of pre-trained LLMs such as GPT-4 Turbo \citep{r2}, GPT-3.5 Turbo, and Llama 2 \citep{r3} raises a pertinent question: \textit{Can these pre-trained LLMs serve as viable substitutes for human experts in evaluating the communication skills of job applicants in HR interviews?} To answer this question, we contribute \textbf{HURIT} (\underline{\textbf{Hu}}man \underline{\textbf{R}}esource \underline{\textbf{I}}nterview \underline{\textbf{T}}ranscripts), a substantial dataset comprising 3,890 transcripts sourced from \textit{Taplingua}\footnote{\textit{Taplingua} is a startup that assists students in preparing for job interviews through its online platform.} and propose the following tasks:

\begin{itemize}
    
    \item \textbf{Task A:} Evaluate the candidate’s performance on several human evaluation criteria individually and \textit{assign scores} on the designated scale.

    \item \textbf{Task B:} \textit{Identify all instances of errors} in the transcript, including disfluency, lack of coherence, tone issues, relevance concerns, verbosity and grammatical inaccuracies. Clearly outline each error to guide the candidate in understanding the areas for improvement.

    \item \textbf{Task C:} \textit{Offer constructive feedback and practical suggestions} for each human evaluation criterion to improve overall performance in HR interview responses.

\end{itemize}


This research seeks to provide a comprehensive analysis of the performance of pre-trained LLMs, including GPT-4 Turbo, GPT-3.5 Turbo, \texttt{text}-\texttt{davinci}-\texttt{003}, \texttt{text}-\texttt{babbage}-\texttt{001} \citep{r4}, \texttt{text}-\texttt{curie}-\texttt{001}, \texttt{text}-\texttt{ada}-\texttt{001}, \texttt{llama}-\texttt{2}-\texttt{7b}-\texttt{chat} \citep{r3}, \texttt{llama}-\texttt{2}-\texttt{13b}-\texttt{chat}, and \texttt{llama}-\texttt{2}-\texttt{70b}-\texttt{chat}. It assesses their effectiveness compared to expert human evaluators in \textit{providing scores}, \textit{pointing out errors}, and \textit{providing feedback and improvement suggestions} to candidates during HR interviews. The ultimate goal is to determine the potential of LLMs in improving and streamlining the evaluation of communication skills for job applicants.


Our findings indicate that GPT-4 Turbo and GPT-3.5 Turbo, in particular, demonstrate commendable performance, \textit{generating evaluation scores} comparable to those of expert human evaluators. Despite their proficiency in \textit{delivering scores} aligned with human evaluators, our in-depth analysis reveals limitations in \textit{pointing out errors}, as well as in the \textit{quality of feedback and improvement suggestions} produced by LLMs compared to human evaluators.

Although LLMs excel in \textit{scoring}, they often lack specificity and actionability when \textit{providing advice} for enhancing candidate performance in HR interviews. This research underscores the need for a human-in-the-loop approach to \textit{detect issues} and provide refined \textit{feedback and improvement suggestions}. We summarize our contributions as follows:

\begin{itemize}
    \item We introduce \textbf{HURIT} (\underline{\textbf{Hu}}man \underline{\textbf{R}}esource \underline{\textbf{I}}nterview \underline{\textbf{T}}ranscripts), a substantial dataset comprising 3,890 transcripts sourced from \textit{Taplingua}, contributing to the field of LLM evaluation in the context of HR interviews.

    
    \item We evaluate the pre-trained LLMs in terms of a set of human evaluation metrics, assessing their effectiveness in comparison to that of human evaluators. This evaluation focuses on \textit{providing scores}, \textit{identifying errors}, and \textit{offering feedback and improvement suggestions} during HR interviews. This paper is \textbf{\textit{the first}} to explore LLMs as an alternative to human evaluation and to show their effectiveness in HR transcript evaluation.

    
\end{itemize}

These contributions collectively demonstrate our commitment to advance the understanding of the capabilities and limitations of pre-trained LLMs in the specific context of HR interview assessment.



\begin{figure*}[h!]
  \centering
     \includegraphics[width=0.96\linewidth]{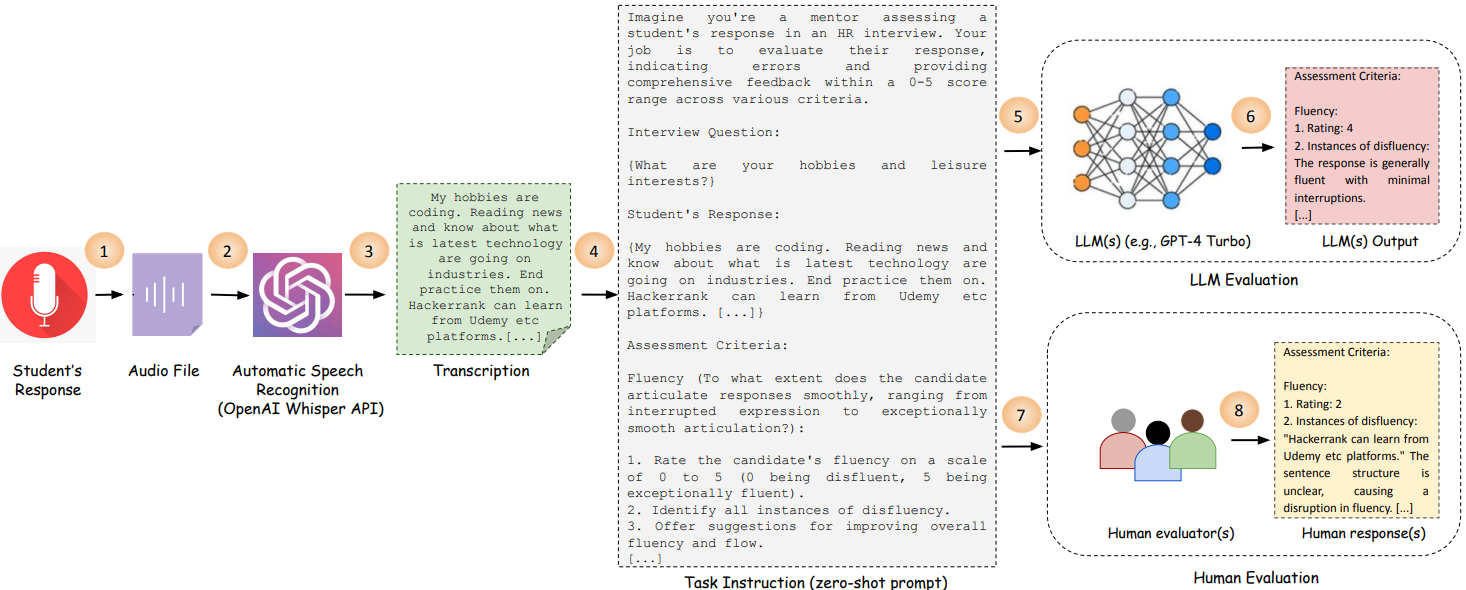}
     \caption{Workflow depicting the process of converting audio transcripts to text using the OpenAI Whisper API, followed by subsequent evaluations using LLMs and human evaluators.} \label{fig1}
\end{figure*}

\section{Related Work}



Earlier research by \citet{r13} delved into feature selection methodologies tailored for automated speech scoring systems, focusing on meeting validity and interpretability criteria while ensuring optimal performance. In a subsequent work, \citet{r14} extracted acoustic and linguistic feature categories using automatic speech recognition (ASR) transcription and audio responses. These features were combined in various configurations to analyze the impact of individual features on performance, particularly for oral proficiency tasks. SpeechRater\textsuperscript{SM} \citep{r12} and other systems use manually engineered features to assess oral proficiency, while recent work, such as that by \citet{r17}, used deep learning techniques such as BiLSTM-Attention to improve scoring results. 

Recent research has also explored specific aspects of speech scoring, such as response content scoring. In this approach, the features extracted from the transcription of the response are modeled alongside the corresponding question to determine the relevance of the response \citep{r18, r19}. Building on this work, \citet{r20} further extended their efforts by incorporating acoustic cues and grammar features to improve scoring performance. In a more recent study, \citet{r21} employed speech and text transformers \citep{r22} to rate candidate speech. Their work includes a notable analysis in which they extract cross-modal information from a response hierarchy, resulting in improved scoring performance.

Automatic scoring systems, crucial for decisions like visas, college admissions, and job interviews, face the challenges of oversensitivity (minor modifications in the transcript lead to significant alterations in the score) and overstability (substantial changes in the transcript cause minimal alterations in the score) \citep{r23}. Studies \citep{r24,r25, r26} have addressed these issues and explored the impact of dataset vulnerabilities. Some approaches involve filtering off-topic responses, improving robustness through reward sampling, and considering institutional trust. Despite these efforts, the black-box nature of many systems limits their interpretability. Although previous analyses focused on feature importance, correlation, and ablation studies, our paper extends this exploration to the specific domain of HR interview evaluations. To our knowledge, no study has explored state-of-the-art LLMs for automatic HR interview assessment. In this paper, we bridge this gap by evaluating prominent pre-trained LLMs on HR interview transcripts, shedding light on their capabilities and limitations in providing feedback to interviewees.

It should be noted that to date, there is no real-world dataset available for HR interview assessment, emphasizing the novelty and challenge of our research in a domain with limited empirical exploration. Furthermore, previous studies in automated scoring focused primarily on \textit{providing scores}, with limited attention to \textit{identifying errors} and \textit{offering feedback and improvement suggestions}. Our research addresses this gap by comprehensively evaluating pre-trained LLMs in the context of HR interview assessments, considering not only \textit{scoring} but also \textit{error identification} and \textit{feedback provision}.

\section{HURIT Dataset}

We collaborated with \textit{Taplingua} (a startup specializing in providing mock HR interviews to students) to collect our dataset. The dataset comprises HR interview transcripts specifically from L2 English speakers, with a focus on interviews from the Asian region. These transcripts are derived from responses in mock HR interviews. The students' responses were recorded in \textit{.mp3} format, and to obtain text transcription, we employed OpenAI's Whisper large-v2 \citep{r1} model. The resulting transcriptions were stored in a structured JSON format with three key fields: "\texttt{id}", "\texttt{name}" and "\texttt{transcription}". The process is shown in steps 1, 2, and 3 of Figure \ref{fig1}. We conducted experiments on four distinct types of questions commonly encountered in HR interviews: 

   1. \textit{\colorbox{blue!10}{Tell me about yourself.}} 
    
   2. \textit{\colorbox{green!10}{What are your strengths and weaknesses?}} 
    
   3. \textit{\colorbox{red!10}{What are your hobbies and leisure interests?}} 
    
   4. \textit{\colorbox{yellow!10}{If you were to meet the CEO of the company,} \colorbox{yellow!10}{what topics would you like to discuss?}} 
    
In total, we collected 3,890 transcriptions from these interview sessions. Table \ref{tab_d} summarizes the key statistics of our proposed dataset. In Table \ref{tab_e}, we provide examples of each type of question along with their corresponding transcriptions.

\begin{table}
\centering
\renewcommand{\arraystretch}{1}
\scalebox{0.62}{
\begin{tabular}{|c|c|c|c|c|}
\hline
\textbf{ Statistic } & \cellcolor{blue!10}Type 1 & \cellcolor{green!10}Type 2 & \cellcolor{red!10}Type 3 & \cellcolor{yellow!10}Type 4 \\
\hline \hline
\# of Transcripts & 1,421 & 1,291 & 989 &  189\\
Avg transcript length (words)  & 70.97 &  76.51 &  72.21 & 68.60 \\
Min transcript length (words) & 20 & 30 &  16 & 22 \\
Max transcript length (words) & 189 & 196 & 152 & 166 \\
\hline
\end{tabular}
}
\caption{\label{tab_d}
Dataset statistics for the HURIT.
}
\end{table}

\begin{table*}
\centering
\renewcommand{\arraystretch}{1}
\scalebox{0.68}{
\begin{tabular}{|c|p{18 cm}|}
\hline
\textbf{Question Type} & \textbf{Example Transcript (Student's Response)}\\
\hline  \hline
 \multirow{2}{*} {\cellcolor{blue!10}1} & Hi, my name is Akash Sen and currently I am pursuing my Mtech in EE from IIT, ISM Dhanbad. Aggregate CGPA of 78 and my hobbies are playing cricket and listening music for relaxation and I graduated in my btech in 2023 from Swami Vivekananda Institute of Science and Technology within aggregate CGPA of 8.41. And apart from this I have done 3 projects in my btech curriculum. First on life, light fidelity, and second is the real time. Real time project in analogue communication which is based on the MATLAB code and other one. \\ \hline

\multirow{2}{*} {\cellcolor{green!10}2} & About my strength.How I was working on my final project. I get to know that my strength was like I can easily attract others. Point so they will accept to my points. I will. I will get all the groups together to work as a team and to be to focus on the work mainly and for my weakness. Words earlier, like during I was not able to manage my time but as I was involving in more projects I get toknow how to do time management and overcome this weakness. \\ \hline

\multirow{2}{*} {\cellcolor{red!10}3} & Hi, my hobbies are, my hobbies are playing cricket, playing cricket, travelling, travelling and volunteering. Volunteering also like volunteering. Also it includes travelling like trekking, trekking also like to do new creative ideas, creative ideas. Creative ideas, including writing, writing, reading, writing, reading,playing, and. Music, paintings etc. \\ \hline

\multirow{2}{*} {\cellcolor{yellow!10}4} & If I met CEO of the company, I would like to discuss about his vision about the future of the company. And I will discuss about my role in this vision, and I will tell him about the what I think about the future of the company. And lastly, I will try to get the feedback about my performance. That's it. \\

\hline
\end{tabular}
}
\caption{\label{tab_e}
Transcript examples of student responses to various types of HR interview questions.
}
\end{table*}

\section{Methodology}

For our comprehensive analysis, we leveraged a diverse set of pre-trained LLMs in zero-shot mode. We use \texttt{GPT}-\texttt{4} \texttt{Turbo} (\texttt{gpt}-\texttt{4}-\texttt{1106}-\texttt{preview}), \texttt{GPT}-\texttt{3.5} \texttt{Turbo} (\texttt{gpt}-\texttt{3.5}-\texttt{turbo}), \texttt{text}-\texttt{davinci}-\texttt{003}, \texttt{text}-\texttt{babbage}-\texttt{001}, \texttt{text}-\texttt{curie}-\texttt{001}, and \texttt{text}-\texttt{ada}-\texttt{001} via the OpenAI API. We also use \texttt{llama}-\texttt{2}-\texttt{7b}-\texttt{chat}, \texttt{llama}-\texttt{2}-\texttt{13b}-\texttt{chat}, and \texttt{llama}-\texttt{2}-\texttt{70b}-\texttt{chat} via the Llama 2-Chat API. We enabled these LLMs to generate responses without specific fine-tuning on HR interview data by prompting these LLMs in their zero-shot configurations.

Our methodology involved a systematic evaluation process, in which both LLMs and human evaluators independently \textit{provided scores}, \textit{identified errors}, and \textit{offered feedback} on HR interview transcripts. This multifaceted approach allowed us to holistically assess the relative performance of each LLM, considering its proficiency in \textit{providing scores}, \textit{identifying errors}, and \textit{offering improvement suggestions}. In addition, we explored their capabilities compared to expert human evaluators across all human evaluation criteria. Each LLM and each human evaluator evaluated all transcripts. 

We hired \textbf{three certified HR managers} through the \href{https://www.upwork.com/}{UpWork} online platform, each with extensive experience conducting HR interviews. These experienced professionals bring their specialized skills to serve as expert evaluators for our HR interview assessment task. 

\subsection{Evaluation Criteria for LLM and Human}

In this section, we present the evaluation criteria used in our study to assess candidate responses in HR interviews. Following \citep{r29, r28} and through consultation with HR managers, these six criteria were identified: 

    \textit{(i) Fluency}: To what extent does the candidate articulate responses smoothly, ranging from interrupted expression to exceptionally smooth articulation?

    \textit{(ii) Coherence}: To what extent is the candidate's response transparent and logically structured?

    \textit{(iii) Tone/Politeness}: How does the candidate's language exhibit the level of formality, respect, and professionalism suitable for an HR interview setting?

    \textit{(iv) Relevance}: How well does the candidate's response directly address and align with the given interview question?

    \textit{(v) Conciseness}: How effectively does the candidate deliver information in a brief, yet informative manner, avoiding unnecessary verbosity?

    \textit{(vi) Grammaticality}: How grammatically correct is the language used in the candidate's response?

\subsection{LLM Evaluation}

The prompt used for the LLM evaluation in this study is shown in Figure \ref{fig2}. We used a single prompt to perform the following tasks and to evaluate all human evaluation criteria.

\noindent \textbf{Task A (\textit{Assign score}):} We zero-shot prompt the pre-trained LLMs to individually evaluate the candidate’s performance for all criteria. The prompt includes explicit instructions to \textit{provide a numerical score} within a defined scale ranging from 0 to 5. This scale serves as a clear metric, with 0 indicating no competence and 5 denoting exceptional competence. 

\noindent \textbf{Task B (\textit{Identify all instances of errors}):}  We zero-shot prompt the pre-trained LLMs to specifically \textit{identify errors} within the transcript related to a given criterion. The prompt clearly outlines the task, focusing on \textit{error identification}.

\noindent \textbf{Task C (\textit{Offer constructive feedback and practical suggestions}):} We zero-shot prompt the pre-trained LLMs to assess the candidate's performance for each criterion. By adhering to this prompt, our objective is to facilitate a holistic evaluation, focusing on overall \textit{feedback and improvement recommendations} for each specific criterion.

The process is depicted in Steps 4, 5, and 6 of Figure \ref{fig1}.
We also explored several other prompting approaches, such as separate prompts for each human evaluation criterion and the few-shot prompting approach, as discussed in more detail in Appendix \ref{app_1a}.

\subsection{Human Evaluation}

To assess the effectiveness of the LLM evaluation results, we compare them with the human evaluation outcomes performed by HR managers. In collaboration with HR managers, we perform \textbf{Task A (\textit{Assign score})}, \textbf{Task B (\textit{Identify all instances of errors})}, and \textbf{Task C (\textit{Offer constructive feedback and practical suggestions})} similar to the LLM evaluation process. For \textbf{Task A}, we calculate the average of the scores assigned by three human evaluators for each human evaluation metric. To ensure a fair and meaningful comparison, we format the prompt in the human evaluation similarly to that used in the LLM evaluation, as shown in Figure \ref{fig2}. The procedure is illustrated in Steps 4, 7, and 8 of Figure \ref{fig1}.

\begin{figure}
    \centering
    \begin{tcolorbox}[enhanced,fit to height=23.5cm,
  colback=violet!20!white!10!white,colframe= violet!65!white,title= Prompt to evaluate spoken transcripts, drop fuzzy shadow]

  {\fontfamily{qcr}\selectfont
 
Imagine you are a mentor assessing a student's response in an HR interview. Your job is to evaluate their response, indicating errors, and providing comprehensive feedback within a 0-5 score range across various criteria. \\

Interview Question:

\{Insert the interview question here\}\\

Student's Response:

\{Insert the interview response transcript here\}\\

Assessment Criteria: \\

Fluency (To what extent does the candidate articulate responses smoothly, ranging from interrupted expression to exceptionally smooth articulation?):

1. Rate the candidate's fluency on a scale of 0 to 5 (0 being disfluent, 5 being exceptionally fluent).

2. Identify all instances of disfluency.

3. Offer suggestions for improving overall fluency and flow.\\

Coherence (To what extent is the
candidate’s response transparent and logically
structured?):

1. Rate the candidate's coherence on a scale of 0 to 5 (0 being incoherent, 5 being highly coherent).

2. Identify all instances where the response lacks coherence.

3. Provide suggestions for improving coherence and maintaining logical flow.\\

Tone/Politeness (How does the candidate’s language exhibit the level of formality, respect, and professionalism suitable for an HR interview setting?):

1. Rate the candidate's tone and politeness on a scale of 0 to 5 (0 representing impolite or unprofessional language, 5 denoting high politeness and professionalism).

2. Identify all instances where the response lacks politeness or professionalism.

3. Offer suggestions for maintaining a respectful and professional tone.\\

Relevance (How well does the candidate’s response directly address and align with the given interview question?):

1. Rate the candidate's relevance to the interview question on a scale of 0 to 5 (0 representing a lack of relevance, 5 indicating a highly relevant response).

2. Identify all off-topic or irrelevant elements in the response.

3. Offer suggestions for maintaining focus and relevance in interview responses.\\

Conciseness (How effectively does the candidate deliver information in a brief, yet informative manner, avoiding unnecessary verbosity?):

1. Rate the candidate's conciseness on a scale of 0 to 5 (0 being overly verbose, 5 being appropriately concise).

2. Identify all instances where the response is excessively wordy or lacks brevity.

3. Offer suggestions for delivering information in a more concise manner.\\

 Grammaticality (How grammatically correct is the language used in the candidate’s response?):

1. Rate the candidate's grammatical accuracy on a scale of 0 to 5 (0 indicating numerous grammatical errors, 5 indicating impeccable grammar).

2. Identify all grammatical errors in the response.

3. Offer suggestions for grammatical improvements.\\

[NOTE: Please provide the evaluation report in the same format as above.]

}
  
\end{tcolorbox}

    \caption{Prompt to evaluate spoken transcripts across six human evaluation criteria.}
    \label{fig2}
\end{figure}

\section{Results}

\begin{table*}
\centering
\renewcommand{\arraystretch}{1}
\scalebox{.73}{
\begin{tabular}{|c|c|c|c|c|c|c|}
\hline
\textbf{Evaluator} & \textbf{Fluency} & \textbf{Coherence} & \textbf{Tone / Politeness} &  \textbf{Relevance} & \textbf{Conciseness} & \textbf{Grammaticality} \\
\hline \hline
{\cellcolor{gray!20}Human} & {\cellcolor{gray!20}2.16} & {\cellcolor{gray!20}2.31} & {\cellcolor{gray!20}3.35} & {\cellcolor{gray!20}2.52} & {\cellcolor{gray!20}2.36} & {\cellcolor{gray!20}2.35}
 \\ \hline \hline
\texttt{text-ada-001} &  3.84 & 4.03 &4.25 &4.15 &4.16 &4.03
\\ 
\texttt{text-curie-001} & 3.16 &3.56 &4.07 &3.91 &3.37 &3.54
 \\
\texttt{text-babbage-001} & 3.50 &4.01 &4.17 &4.01 &4.00 &3.81
 \\ 
\texttt{text-davinci-003} &  2.94 &3.21 &3.93 &3.49 &2.94 &3.27
\\
\texttt{GPT-3.5 Turbo} & 2.49 &1.93 &3.31 &2.84 &2.04 &2.23
 \\ 
\texttt{GPT-4 Turbo} & 2.44 &1.98 &3.38 &2.79 &2.23 &2.17
 \\
\texttt{llama-2-7b-chat}&  3.21 &3.79 &4.21 &3.50 &3.92 &3.78
 \\ 
\texttt{llama-2-13b-chat} &  2.92 &3.59 &3.99 &3.69 &3.11 &3.46
\\
\texttt{llama-2-70b-chat} & 2.72 &3.26 &3.75 &3.46 &2.94 &3.24
\\ 
\hline
\end{tabular}
}
\caption{\label{tab_r}
Comparison of average assigned scores for all speech transcripts across four question types between human evaluators and LLMs based on six human evaluation criteria.
}
\end{table*}

\begin{table}
\centering
\renewcommand{\arraystretch}{1}
\scalebox{0.60}{
\begin{tabular}{|c|c|c|}
\hline
\textbf{ Criterion } & \textbf{GPT-3.5 Turbo} & \textbf{GPT-4 Turbo} \\
\hline \hline
Fluency & 0.08 & 0.21 \\
Coherence & 0.07 & 0.19\\
Tone / Politeness & 0.38 & 0.44 \\ 
Relevance & 0.18 & 0.23 \\ 
Conciseness & 0.15 & 0.27 \\
Grammaticality & 0.18 & 0.22
\\
\hline  
\end{tabular}
}
\caption{\label{tab_pe}
The Kendall’s correlation coefficient ($\tau$) between best-performing LLMs (i.e., GPT-4 Turbo and GPT-3.5 Turbo) and human evaluators.
}
\end{table}

\subsection{Do LLM and human evaluators agree on how they \textit{score} different aspects of communication? (\textit{Task A})}

In evaluating the alignment of LLMs with human evaluators across various communication criteria, Table \ref{tab_r} offers a comprehensive view. In particular, GPT-4 Turbo and GPT-3.5 Turbo consistently demonstrate close agreement with human evaluations, showcasing their superior performance across multiple facets of communication, including fluency, coherence, tone/politeness, relevance, conciseness, and grammaticality. \texttt{Llama}-\texttt{2}-\texttt{13b}-\texttt{chat} and \texttt{llama}-\texttt{2}-\texttt{70b}-\texttt{chat} also exhibit notable agreement with human assessments, though not as consistently
as GPT-4 Turbo and GPT-3.5 Turbo. Other LLMs, such as \texttt{text}-\texttt{ada}-\texttt{001}, \texttt{text}-\texttt{curie}-\texttt{001}, \texttt{text}-\texttt{babbage}-\texttt{001}, \texttt{text}-\texttt{davinci}-\texttt{003}, and \texttt{llama}-\texttt{2}-\texttt{7b}-\texttt{chat}, show varied degrees of performance across the evaluated criteria. Although they exhibit diversity in their performance, they are not consistently better than GPT-3.5 Turbo or GPT-4 Turbo. We also discuss the performance based on each question type in assigning scores in Appendix \ref{app_b1}.


Following \citet{r31}, we also calculated the Kendall’s correlation coefficient ($\tau$) between the scores assigned by these best-performing GPT models (i.e., GPT-3.5 Turbo and GPT-4 Turbo) and the assessments provided by HR managers across various communication aspects. In particular, as seen in Table \ref{tab_pe}\footnote{When interpreting Kendall’s $\tau$, $|\tau| \in [0,0.1)$ is considered as very weak correlation, $|\tau| \in [0.1,0.2)$ is considered as weak correlation, $|\tau| \in [0.2,0.3)$ is considered as moderate correlation and $|\tau| \in [0.3,1.0]$ is considered as strong correlation \citep{r27}.}, the analysis reveals nuanced insights into the alignment between LLMs and HR managers' judgments in the context of HR interview evaluations. In terms of fluency, GPT-3.5 Turbo shows a very weak correlation, while GPT-4 Turbo exhibits a moderate correlation with HR managers' assessments. Moving to coherence, GPT-3.5 Turbo shows a very weak correlation, whereas GPT-4 Turbo indicates a weak correlation. Transitioning to tone/politeness, both GPT-3.5 Turbo and GPT-4 Turbo demonstrate a strong correlation. In terms of relevance, conciseness, and grammaticality, GPT-3.5 Turbo shows a weak correlation, while GPT-4 Turbo exhibits a moderate correlation. It is important to note that we observed very weak, weak, or moderate correlations for all criteria between HR managers and GPT models (i.e., GPT-3.5 Turbo and GPT-4 Turbo), except for tone/politeness. This indicates room for further refinement to align the LLMs with the HR managers’ assessments.

To assess consensus among the three HR managers in assigning scores to each spoken transcript for each human evaluation criterion, we employ Fleiss’s kappa as a measure of inter-annotator agreement. Our computations reveal agreement scores of 0.52, 0.59, 0.48, 0.43, 0.41, and 0.60 for fluency, coherence, tone/politeness, relevance, conciseness, and grammaticality. The kappa values for these criteria show a moderate level of agreement \citep{r30}.

\subsection{How does the LLM evaluation compare to the human evaluation in \textit{identifying errors}, \textit{issues}, or \textit{inconsistencies} in the candidate's communication? (\textit{Task B})}

As we have seen, GPT-3.5 Turbo and GPT-4 Turbo showed better results in assigning scores to spoken transcripts compared to other LLMs. We compared various evaluations provided by GPT-3.5 Turbo, GPT-4 Turbo, and human evaluators, and the exact evaluations are provided in Appendix \ref{app2}. 

GPT-3.5 Turbo superficially discusses the various fluency issues and interruptions in the interview transcripts. GPT-4 Turbo delves into fluency issues in interview transcripts in more detail, citing various examples of disfluency. There is also a problem of repetition of ideas that GPT-3.5 Turbo and GPT-4 Turbo are not able to capture properly. Sometimes, the transition between sentences is not fluent. Unlike GPT-3.5 Turbo and GPT-4 Turbo, human
evaluators are better at identifying interruptions
and repetitions in the transcript that contribute to
disfluency.

While GPT-3.5 Turbo identifies sentences lacking coherence due to run-on sentences and grammatical issues, and GPT-4 Turbo can detect the jumbling of various ideas, neither can pinpoint all of these issues effectively. In contrast, human evaluators excel at correctly identifying all incoherence issues in interview transcripts, emphasizing the absence of a smooth transition between two separate ideas. Human evaluators can also properly locate the disjointed structure of the spoken transcript, leading to incoherence. Sometimes, the speaker is not able to correlate between ideas and suddenly jumps from one idea to another without any interconnection between the ideas. In such cases, unlike GPT-3.5 Turbo or GPT-4 Turbo, human evaluators can point out that there are instances of multitasking without a proper transition from one idea to the next, which leads to incoherence.

GPT-4 Turbo and GPT-3.5 Turbo are mostly able to detect politeness or professionalism issues in interview transcripts closely on par with human evaluators. GPT-4 Turbo performs better than GPT-3.5 Turbo in discussing any politeness issues in the interview transcripts.

The idea behind the relevance between human evaluators and these LLMs evaluations also differs to some extent. For instance, according to GPT-4 Turbo and GPT-3.5 Turbo, a response has irrelevant parts.  However, according to the human evaluator, the response is mostly relevant. Furthermore, there are places where these LLMs and human evaluators agree that there are certain portions of the response that lack relevance to the context of the question. Human evaluators can better understand and accurately highlight irrelevant issues in the transcripts, as compared to these GPT models.

Both GPT-4 Turbo and GPT-3.5 Turbo are unable to pinpoint the areas of response that result in longer than required responses. However, human evaluators can identify redundant portions in student responses that can be improved for a more concise representation. Sometimes, the responses are verbose and lack a clear structure. Unlike LLMs, human evaluators excel at detecting unnecessary repetitions and pinpointing them. 

Human evaluators are also better than GPT-4 Turbo and GPT-3.5 Turbo in finding grammatical flaws in the student responses. Human evaluators point out grammatical flaws in the response, while GPT-4 Turbo either partially misses such flaws or provides a generalized evaluation without explicitly stating them. GPT-3.5 Turbo is not very good at pointing out the grammatical flaws, as mostly it only provides a generalized evaluation without explicitly stating them. Unlike LLMs, human evaluators are better at detecting various grammatical errors such as incorrect verb conjugation, missing conjunctions, incorrect sentence structure, etc.

In general, we can say that these LLMs have limitations in fully grasping and addressing errors and issues. Although their evaluations capture some aspects of awkward phrasing, incoherence, grammaticality, politeness, conciseness, and relevance, the nuances identified by the human evaluator go beyond the capabilities of LLMs. Unlike these LLMs, human evaluator can pinpoint errors, issues, and inconsistencies in student responses. Therefore, LLMs have not yet been able to reach the level of human performance to find flaws in interview transcripts. In particular, GPT-3.5 Turbo lags far behind human evaluators and GPT-4 Turbo does not fully meet human performance standards.

\subsection{How do the LLM and humans compare in terms of \textit{providing suggestions and feedback} to improve communication skills? (\textit{Task C})}

As highlighted earlier, GPT-3.5 Turbo and GPT-4 Turbo demonstrated superior performance in assigning scores to spoken transcripts compared to other LLMs. In this analysis,
we examine various evaluations generated by GPT-3.5 Turbo, GPT-4 Turbo, and human evaluators, and the specific evaluations are detailed in Appendix \ref{app2}. 

In terms of providing suggestions for fluency, GPT-3.5 Turbo generally gives superficial suggestions to improve the fluency of the response. GPT-4 Turbo is slightly better than GPT-3.5 Turbo in describing exactly what ideas are needed to improve the response. Unlike these GPT models, human evaluators excel in providing precise suggestions, including recommending rephrased versions of sentences. Moreover, they emphasize speaking in a structured manner and using varied sentence constructions to avoid repetition. 


For coherence, there are certain contradictions between the suggestions provided by GPT-4 Turbo and GPT-3.5 Turbo compared to human evaluators. Human evaluators have given more structured and logical feedback, which, when implemented, can lead to significant improvements in the students' responses. Human evaluators have also provided suggestions for rephrasing some sentences to make them coherent. LLMs like GPT-3.5 Turbo, on the other hand, give feedback responses that contradict human feedback. GPT-4 Turbo is better than GPT-3.5 Turbo in providing improvement suggestions, but still lacks clarity in the suggestions. 

In terms of suggesting improvements for tone/politeness, GPT-4 Turbo performs almost on par with human evaluators, although GPT-3.5 Turbo lags behind human evaluators in providing critical suggestions to improve the tone of the response. 

Human evaluators are also better at measuring the relevance of transcripts compared to GPT-4 Turbo and GPT-3.5 Turbo. Human evaluators clearly understand the context of the response and provide fruitful suggestions to improve the relevance of the response. GPT-4 Turbo and GPT-3.5 Turbo, on the other hand, sometimes provide suggestions and feedback that might be irrelevant or inappropriate in certain cases. 

In terms of conciseness, the feedback provided by GPT-4 Turbo sometimes lacks key details on how to make the response more concise. In contrast, human evaluators provide clear suggestions, such as rephrasing sentences, removing repetitive phrases, and presenting certain points concisely.

Human evaluators are better at finding grammatical flaws in the response and providing clear suggestions to improve them. LLMs like GPT-4 Turbo and GPT-3.5 Turbo are not able to always find all the possible grammatical issues in the response, thus not being able to provide improvement suggestions effectively as compared to human evaluators. Human evaluators also provide good alternatives on how to restate certain portions of the response to improve upon the grammatical flow of the sentences. LLMs, though providing decent suggestions in terms of grammatical improvement, are still not able to match the level of human evaluators.

 Although GPT-4 Turbo outperforms GPT-3.5 Turbo in some aspects, both GPT-3.5 Turbo and GPT-4 Turbo fail to provide feedback that matches the precision and clarity of human evaluators. Human evaluators excel in offering nuanced suggestions for fluency, relevance, and conciseness, as well as demonstrating a superior ability to address grammatical flaws and improve overall coherence. Despite advancements in LLMs, they still struggle to rival the comprehensive understanding and insightful feedback provided by human evaluators across various dimensions of quality in spoken interview transcripts.

\section{Conclusion}

Our work introduces the HURIT dataset, a valuable compilation of HR interview spoken transcripts. Our thorough analysis focuses on assessing the performance of pre-trained LLMs in evaluating spoken HR interview transcripts. These LLMs are systematically compared to expert human evaluators to gauge their effectiveness in \textit{providing scores}, \textit{identifying errors}, and \textit{offering feedback} during simulated HR interviews. Our findings emphasize the commendable performance of pre-trained LLMs, particularly highlighting the success of GPT-4 Turbo and GPT-3.5 Turbo, which demonstrate evaluations comparable to those provided by expert human evaluators. However, despite their proficiency in \textit{scoring} aligned with human evaluation metrics, these LLMs fall short of human evaluators in accurately \textit{identifying errors} and \textit{delivering specific, actionable advice} for candidate improvement in HR interviews. This limitation underscores the challenges that pre-trained LLMs currently face in fully replicating the nuanced understanding and feedback capabilities demonstrated by human experts in the HR interview context. This research suggests that the current state-of-the-art pre-trained LLMs may not be fully ready for automated deployment in HR interview assessments. Instead, we propose a human-in-the-loop approach, emphasizing manual checks for inconsistencies and provisions to enhance feedback quality. The identified room for improvement highlights the ongoing need for research and development to refine the capabilities of LLMs in the context of HR interview evaluations.

\section*{Limitations}

The study explores only four types of HR interview questions. The findings may not be generalizable to a broader range of interview questions, and future research should consider including a more diverse set of question types to enhance the comprehensiveness of the analysis. The HR spoken transcripts used in this study are exclusively from the Asian region. As a result, the findings may not be representative of global HR interview practices and caution should be exercised when applying the results to different cultural and regional contexts. Future research should aim to collect data from a more geographically diverse sample of candidates to improve the external validity of the study.

\section*{Ethics Statement}

We declare that the dataset used in this article has been used solely for research and educational purposes, without commercial exploitation. Moreover, the identities of the candidates (e.g., name, affiliation, grades, etc.) in the spoken transcripts have been changed to protect the privacy of the concerned candidates. Our utilization of GPT models adheres strictly to the \href{https://openai.com/policies/usage-policies}{OpenAI usage policy}, ensuring compliance with the stipulated guidelines. Furthermore, all necessary \href{https://openai.com/pricing}{charges} for the use of the OpenAI API have been duly paid, in accordance with their terms and conditions. In the recruitment of human evaluators, we have upheld the principles of fairness and equity by compensating them at rates consistent with those outlined by \href{https://www.upwork.com/legal}{UpWork}. All experiments conducted throughout this study adhere strictly to the guidelines established by the \href{https://www.aclweb.org/portal/content/acl-code-ethics}{ACL Code of Ethics}.





\bibliography{anthology,custom}

\appendix
\label{sec:appendix}


\section{Methodology} \label{app_1a}

\subsection{Separate prompts for each human evaluation criterion}

In our methodology, we try zero-shot prompting on LLMs with individual prompts for each human evaluation criterion, including \textit{fluency} (Figure \ref{fig_sp1}), \textit{coherence} (Figure \ref{fig_sp2}), \textit{tone/politeness} (Figure \ref{fig_sp3}), \textit{relevance} (Figure \ref{fig_sp4}), \textit{conciseness} (Figure \ref{fig_sp5}), and \textit{grammaticality} (Figure \ref{fig_sp6}). For each criterion, the instruction included the interview question, the student's response, and instructions for three distinct tasks: \textit{\textbf{assigning scores}} (\textbf{Task A}), \textit{\textbf{identifying errors}} (\textbf{Task B}), and \textit{\textbf{providing constructive feedback with practical suggestions}} (\textbf{Task C}). Despite our thorough approach, the results did not reveal improvements in all the criteria evaluated. Consequently, results derived from this zero-shot prompt approach for each criterion are not included in this paper.

\subsection{Few-shot prompting}

We also explore the few-shot prompt technique, incorporating a prompt that evaluates all criteria at once, as illustrated in Figure \ref{fig2}. Few-shot examples were explored in our experiments, which included the interview question, the student's response, and evaluation results for each criterion. The exploration encompassed \texttt{two-shot}, \texttt{four-shot}, and \texttt{eight-shot} examples. We specifically investigated the few-shot prompting approach with GPT-3.5 Turbo and GPT-4 Turbo, as they demonstrated the best performance among LLMs. Despite our methodical approach, the results obtained through few-shot prompting indicated no substantial enhancements across the evaluated criteria. Consequently, the findings derived from the few-shot prompting approach are not incorporated in this paper. 

\begin{figure*}
    \centering
    \begin{tcolorbox}[enhanced,fit to height=7cm,
  colback=red!20!white!10!white,colframe= red!55!white,title= Prompt to evaluate \textit{fluency} in spoken transcripts, drop fuzzy shadow]

  {\fontfamily{qcr}\selectfont
 
Imagine you are a mentor assessing a student's response in an HR interview. Your job is to evaluate their response, indicating errors, and providing comprehensive feedback within a 0-5 score range for the fluency assessment criterion. \\

Interview Question:

\{Insert the interview question here\}\\

Student's Response:

\{Insert the interview response transcript here\}\\

Assessment Criterion: \\

Fluency (To what extent does the candidate articulate responses smoothly, ranging from interrupted expression to exceptionally smooth articulation?):

1. Rate the candidate's fluency on a scale of 0 to 5 (0 being disfluent, 5 being exceptionally fluent).

2. Identify all instances of disfluency.

3. Offer suggestions for improving overall fluency and flow.\\

[NOTE: Please provide the evaluation report in the same format as above.]

}
  
\end{tcolorbox}

    \caption{ Prompt to evaluate \textit{fluency} in spoken transcripts.}
    \label{fig_sp1}
\end{figure*}

\begin{figure*}
    \centering
    \begin{tcolorbox}[enhanced,fit to height=7cm,
  colback=red!20!white!10!white,colframe= red!55!white,title=Prompt to evaluate \textit{coherence} in spoken transcripts, drop fuzzy shadow]

  {\fontfamily{qcr}\selectfont
 
Imagine you are a mentor assessing a student's response in an HR interview. Your job is to evaluate their response, indicating errors, and providing comprehensive feedback within a 0-5 score range for the coherence assessment criterion. \\

Interview Question:

\{Insert the interview question here\}\\

Student's Response:

\{Insert the interview response transcript here\}\\

Assessment Criterion: \\

Coherence (To what extent is the
candidate’s response transparent and logically
structured?):

1. Rate the candidate's coherence on a scale of 0 to 5 (0 being incoherent, 5 being highly coherent).

2. Identify all instances where the response lacks coherence.

3. Provide suggestions for improving coherence and maintaining logical flow.\\

[NOTE: Please provide the evaluation report in the same format as above.]

}
  
\end{tcolorbox}

    \caption{Prompt to evaluate \textit{coherence} in spoken transcripts.}
    \label{fig_sp2}
\end{figure*}

\begin{figure*}
    \centering
    \begin{tcolorbox}[enhanced,fit to height=7cm,
  colback=red!20!white!10!white,colframe= red!55!white,title=Prompt to evaluate \textit{tone/politeness} in spoken transcripts, drop fuzzy shadow]

  {\fontfamily{qcr}\selectfont
 
Imagine you are a mentor assessing a student's response in an HR interview. Your job is to evaluate their response, indicating errors, and providing comprehensive feedback within a 0-5 score range for the tone/politeness assessment criterion. \\

Interview Question:

\{Insert the interview question here\}\\

Student's Response:

\{Insert the interview response transcript here\}\\

Assessment Criterion: \\

Tone/Politeness (How does the candidate’s language exhibit the level of formality, respect, and professionalism suitable for an HR interview setting?):

1. Rate the candidate's tone and politeness on a scale of 0 to 5 (0 representing impolite or unprofessional language, 5 denoting high politeness and professionalism).

2. Identify all instances where the response lacks politeness or professionalism.

3. Offer suggestions for maintaining a respectful and professional tone.\\

[NOTE: Please provide the evaluation report in the same format as above.]

}
  
\end{tcolorbox}

    \caption{Prompt to evaluate \textit{tone/politeness} in spoken transcripts.}
    \label{fig_sp3}
\end{figure*}

\begin{figure*}
    \centering
    \begin{tcolorbox}[enhanced,fit to height=7cm,
  colback=red!20!white!10!white,colframe= red!55!white,title=Prompt to evaluate \textit{relevance} in spoken transcripts, drop fuzzy shadow]

  {\fontfamily{qcr}\selectfont
 
Imagine you are a mentor assessing a student's response in an HR interview. Your job is to evaluate their response, indicating errors, and providing comprehensive feedback within a 0-5 score range for the relevance assessment criterion. \\

Interview Question:

\{Insert the interview question here\}\\

Student's Response:

\{Insert the interview response transcript here\}\\

Assessment Criterion: \\

Relevance (How well does the candidate’s response directly address and align with the given interview question?):

1. Rate the candidate's relevance to the interview question on a scale of 0 to 5 (0 representing a lack of relevance, 5 indicating a highly relevant response).

2. Identify all off-topic or irrelevant elements in the response.

3. Offer suggestions for maintaining focus and relevance in interview responses.\\

[NOTE: Please provide the evaluation report in the same format as above.]

}
  
\end{tcolorbox}

    \caption{Prompt to evaluate \textit{relevance} in spoken transcripts.}
    \label{fig_sp4}
\end{figure*}

\begin{figure*}
    \centering
    \begin{tcolorbox}[enhanced,fit to height=7cm,
  colback=red!20!white!10!white,colframe= red!55!white,title=Prompt to evaluate \textit{conciseness} in spoken transcripts, drop fuzzy shadow]

  {\fontfamily{qcr}\selectfont
 
Imagine you are a mentor assessing a student's response in an HR interview. Your job is to evaluate their response, indicating errors, and providing comprehensive feedback within a 0-5 score range for the conciseness assessment criterion. \\

Interview Question:

\{Insert the interview question here\}\\

Student's Response:

\{Insert the interview response transcript here\}\\

Assessment Criterion: \\

Conciseness (How effectively does the candidate deliver information in a brief, yet informative manner, avoiding unnecessary verbosity?):

1. Rate the candidate's conciseness on a scale of 0 to 5 (0 being overly verbose, 5 being appropriately concise).

2. Identify all instances where the response is excessively wordy or lacks brevity.

3. Offer suggestions for delivering information in a more concise manner.\\

[NOTE: Please provide the evaluation report in the same format as above.]

}
  
\end{tcolorbox}

    \caption{Prompt to evaluate \textit{conciseness} in spoken transcripts.}
    \label{fig_sp5}
\end{figure*}

\begin{figure*}
    \centering
    \begin{tcolorbox}[enhanced,fit to height=7cm,
  colback=red!20!white!10!white,colframe= red!55!white,title=Prompt to evaluate \textit{grammaticality} in spoken transcripts, drop fuzzy shadow]

  {\fontfamily{qcr}\selectfont
 
Imagine you are a mentor assessing a student's response in an HR interview. Your job is to evaluate their response, indicating errors, and providing comprehensive feedback within a 0-5 score range for the grammaticality assessment criterion. \\

Interview Question:

\{Insert the interview question here\}\\

Student's Response:

\{Insert the interview response transcript here\}\\

Assessment Criterion: \\

 Grammaticality (How grammatically correct is the language used in the candidate’s response?):

1. Rate the candidate's grammatical accuracy on a scale of 0 to 5 (0 indicating numerous grammatical errors, 5 indicating impeccable grammar).

2. Identify all grammatical errors in the response.

3. Offer suggestions for grammatical improvements.\\

[NOTE: Please provide the evaluation report in the same format as above.]

}
  
\end{tcolorbox}

    \caption{Prompt to evaluate \textit{grammaticality} in spoken transcripts.}
    \label{fig_sp6}
\end{figure*}

\section{Results}

\subsection{Analyzing the performance of various LLMs in comparison to human evaluation in \textit{assigning scores} (\textit{Task A}) to various aspects of communication for each question type} \label{app_b1}

In this section, we discuss the performance of various LLMs in \textit{\underline{assigning scores}} (\textit{\textbf{Task A}}) to different human evaluation metrics and compare them to human baselines. The comparison is shown in Table \ref{tab_r1} and is \underline{based on four question types}. Each type of question, namely:

   1. \textit{\colorbox{blue!10}{Tell me about yourself.}} 
    
   2. \textit{\colorbox{green!10}{What are your strengths and weaknesses?}} 
    
   3. \textit{\colorbox{red!10}{What are your hobbies and leisure interests?}} 
    
   4. \textit{\colorbox{yellow!10}{If you were to meet the CEO of the company,} \colorbox{yellow!10}{what topics would you like to discuss?}} 
   
   has been analyzed with a focus on six human evaluation criteria: \textit{fluency}, \textit{coherence}, \textit{tone/politeness}, \textit{relevance}, \textit{conciseness}, and \textit{grammaticality}. The goal is to identify which LLMs perform most closely with human-assigned scores, providing valuable insights into the effectiveness of these LLMs in evaluating spoken responses.

\subsubsection{Type 1: \textit{\colorbox{blue!10}{Tell me about yourself.}}}

In analyzing the performance of various LLMs for the first question type, "\textit{Tell me about yourself.}", concerning the assignment of scores to different human evaluation metrics (\textbf{\textit{Task A}}), distinct patterns emerged. 

In particular, \texttt{text}-\texttt{ada}-\texttt{001} exhibited the most substantial deviation from human evaluations, followed by \texttt{text}-\texttt{babbage}-\texttt{001} and \texttt{llama}-\texttt{2}-\texttt{7b}-\texttt{chat}. Subsequently, \texttt{text}-\texttt{curie}-\texttt{001}, \texttt{text}-\texttt{davinci}-\texttt{003}, \texttt{llama}-\texttt{2}-\texttt{13b}-\texttt{chat}, and \texttt{llama}-\texttt{2}-\texttt{70b}-\texttt{chat} demonstrated intermediate performance. Among LLMs, GPT-3.5 Turbo consistently demonstrated improved proficiency, while GPT-4 Turbo emerged as the closest performer to human assessments. These findings underscore the superior performance of GPT-3.5 Turbo and GPT-4 Turbo in accurately assigning scores for responses to the Type 1 question, highlighting their effectiveness in this conversational context.

\subsubsection{Type 2: \textit{\colorbox{green!10}{What are your strengths and} \colorbox{green!10}{weaknesses?}}}

In evaluating the performance of various LLMs for the second question type, "\textit{What are your strengths and weaknesses?}" with a focus on the assignment of scores to different human evaluation metrics (\textbf{\textit{Task A}}), a discernible pattern emerged. 

Notably, \texttt{text}-\texttt{ada}-\texttt{001} displayed the most significant deviation from human evaluations, followed by \texttt{text}-\texttt{babbage}-\texttt{001} and \texttt{text}-\texttt{curie}-\texttt{001}. Subsequently, \texttt{llama}-\texttt{2}-\texttt{7b}-\texttt{chat}, \texttt{llama}-\texttt{2}-\texttt{13b}-\texttt{chat}, \texttt{text}-\texttt{davinci}-\texttt{003}, and \texttt{llama}-\texttt{2}-\texttt{70b}-\texttt{chat} demonstrated intermediate performance. Among LLMs, GPT-3.5 Turbo consistently exhibited improved proficiency, while GPT-4 Turbo emerged as the closest performer to human assessments. These results underscore the superior performance of GPT-3.5 Turbo and GPT-4 Turbo in accurately assigning scores for responses to the Type 2 question, highlighting their effectiveness in this conversational context.

\subsubsection{Type 3: \textit{\colorbox{red!10}{What are your hobbies and} \colorbox{red!10}{leisure interests?}}}

In assessing the performance of various LLMs for the third question type, "\textit{What are your hobbies and leisure interests?}" with a focus on the assignment of scores to different human evaluation metrics (\textbf{\textit{Task A}}), a distinctive trend emerged. 

Notably, \texttt{text}-\texttt{ada}-\texttt{001} exhibited the most substantial deviation from human evaluations, followed by \texttt{text}-\texttt{babbage}-\texttt{001} and \texttt{llama}-\texttt{2}-\texttt{7b}-\texttt{chat}. Subsequently, \texttt{text}-\texttt{curie}-\texttt{001}, \texttt{llama}-\texttt{2}-\texttt{13b}-\texttt{chat}, \texttt{text}-\texttt{davinci}-\texttt{003}, and \texttt{llama}-\texttt{2}-\texttt{70b}-\texttt{chat} demonstrated intermediate performance. Among LLMs, GPT-3.5 Turbo consistently demonstrated improved proficiency, while GPT-4 Turbo emerged as the closest performer to human assessments. These findings underscore the superior performance of GPT-3.5 Turbo and GPT-4 Turbo in accurately assigning scores for responses to the Type 3 question, emphasizing their effectiveness in this conversational context.

\subsubsection{Type 4: \textit{\colorbox{yellow!10}{If you were to meet the CEO of} \colorbox{yellow!10}{the company, what topics would you like} \colorbox{yellow!10}{to discuss?}}}

In evaluating the performance of various LLMs for the fourth type of question "\textit{If you were to meet the CEO of the company, what topics would you like to discuss?}" with a focus on the assignment of scores to different human evaluation metrics (\textbf{\textit{Task A}}), a discernible pattern emerged. 

In particular, \texttt{text}-\texttt{ada}-\texttt{001} exhibited the most substantial deviation from human evaluations, followed by \texttt{llama}-\texttt{2}-\texttt{7b}-\texttt{chat}, \texttt{text}-\texttt{babbage}-\texttt{001}, and \texttt{text}-\texttt{curie}-\texttt{001}. Subsequently, \texttt{text}-\texttt{davinci}-\texttt{003}, \texttt{llama}-\texttt{2}-\texttt{13b}-\texttt{chat}, and \texttt{llama}-\texttt{2}-\texttt{70b}-\texttt{chat} demonstrated intermediate performance. Among LLMs, GPT-3.5 Turbo consistently demonstrated improved proficiency, while GPT-4 Turbo emerged as the closest performer to human assessments. These results highlight the superior performance of GPT-3.5 Turbo and GPT-4 Turbo in accurately assigning scores for responses to the Type 4 question, underscoring their effectiveness in this conversational context.

Our comprehensive analysis of different LLMs to \textit{assign scores} for four different question types provides valuable insights into their performance across various human evaluation metrics. In particular, \texttt{text}-\texttt{ada}-\texttt{001} consistently deviated the most from human evaluations, indicating challenges in accurately assessing responses. On the contrary, GPT-3.5 Turbo and GPT-4 Turbo consistently emerged as the top performers, showcasing their remarkable ability to align closely with human assessments. These findings underscore the robustness and effectiveness of GPT-3.5 Turbo and GPT-4 Turbo in capturing the nuances of diverse conversational scenarios. 

\begin{table*}
\centering
\renewcommand{\arraystretch}{1}
\scalebox{0.75}{
\begin{tabular}{|c|c|c|c|c|c|c|}
\hline
\textbf{Evaluator} & \textbf{Fluency} & \textbf{Coherence} & \textbf{Tone / Politeness} &  \textbf{Relevance} & \textbf{Conciseness} & \textbf{Grammaticality} \\
\hline \hline
\multicolumn{7}{|c|}{\textit{{\cellcolor{blue!10}Type 1: Tell me about yourself.}}} \\ \hline \hline
{\cellcolor{gray!20}Human} & \cellcolor{gray!20}2.24 & \cellcolor{gray!20}2.50 & \cellcolor{gray!20}3.54 & \cellcolor{gray!20}2.64 & \cellcolor{gray!20}2.56 & \cellcolor{gray!20}2.59
 \\ \hline \hline
\texttt{text-ada-001} &  4.11 & 3.79 &3.89 &4.18 &3.69 &3.76
 \\ 
\texttt{text-curie-001} &  3.10 & 3.72 &4.01 &3.82 &3.77 &3.42
 \\
\texttt{text-babbage-001} &  3.66 & 4.06 &3.86 &3.99 &3.80 &3.59
 \\ 
\texttt{text-davinci-003} &  2.89 &3.50 &4.13 &3.29 &2.99 &3.25
 \\
\texttt{GPT-3.5 Turbo} &  2.66 &2.03 &3.84 &3.25 &2.10 &3.10
 \\ 
\texttt{GPT-4 Turbo} &  2.62 &2.11 &3.79 &3.23 &2.29 &3.02
 \\
\texttt{llama-2-7b-chat}&  3.11 &3.78 &4.02 & 3.94 &3.89 &3.79
 \\ 
\texttt{llama-2-13b-chat} &  2.89 & 3.49 & 3.99 &3.49 &3.24 &3.45
 \\
\texttt{llama-2-70b-chat} &  2.69 & 3.39 &3.89 &3.19 &2.89 &3.14
 \\ 
\hline \hline 
\multicolumn{7}{|c|}{\textit{\cellcolor{green!10} Type 2: What are your strengths and weaknesses?}} \\ \hline \hline
\cellcolor{gray!20}Human &  \cellcolor{gray!20}2.10 & \cellcolor{gray!20}2.25 & \cellcolor{gray!20}3.12 & \cellcolor{gray!20}2.43 & \cellcolor{gray!20}2.24 & \cellcolor{gray!20}1.96
 \\ \hline \hline
\texttt{text-ada-001} & 3.67 &4.17 &4.15 &4.04 &4.30 &4.21
 \\ 
\texttt{text-curie-001} & 3.01 & 3.13 &3.76 &3.99 &3.13 &3.67
 \\
\texttt{text-babbage-001} & 3.27 & 3.93 &3.99 &3.97 &3.88 &4.17
 \\ 
\texttt{text-davinci-003} & 2.89 & 2.79 &3.59 &3.59 &3.09 &3.09
 \\
\texttt{GPT-3.5 Turbo} & 2.99 &1.89 &2.80 &3.11 &2.00 &1.42
 \\ 
\texttt{GPT-4 Turbo} & 2.89 &1.89 &2.89 &2.99 &2.19 &1.39
 \\
\texttt{llama-2-7b-chat}& 2.89 &3.79 &3.87 &3.84 &3.39 &3.59
 \\ 
\texttt{llama-2-13b-chat} &  2.74 &3.74 & 3.69 &3.79 &3.01 &3.44
 \\
\texttt{llama-2-70b-chat} &  2.69 & 3.34 &3.49 &3.49 &2.99 &3.19
 \\ 
\hline \hline
\multicolumn{7}{|c|}{\textit{\cellcolor{red!10} Type 3: What are your hobbies and leisure interests?}} \\ \hline \hline
\cellcolor{gray!20}Human & \cellcolor{gray!20}2.02 & \cellcolor{gray!20}2.34 & \cellcolor{gray!20}3.41 & \cellcolor{gray!20}2.32 & \cellcolor{gray!20}2.20 & \cellcolor{gray!20}2.43
 \\ \hline \hline
\texttt{text-ada-001} &3.70 &4.13 &4.84 &4.26 &4.67 &4.08
 \\ 
\texttt{text-curie-001} & 3.38 &3.77 &4.56 &3.98 &3.14 &3.51
 \\
\texttt{text-babbage-001} & 3.54 &3.98 &4.88 &4.07 &3.98 &3.65
 \\ 
\texttt{text-davinci-003} & 2.98 &3.18 &4.08 &3.78 &2.68 &3.48
 \\
\texttt{GPT-3.5 Turbo} &  1.71 & 1.71 &3.22 &1.69 &1.89 &2.09
 \\ 
\texttt{GPT-4 Turbo} & 1.68 &1.78 &3.48 &1.68 &2.18 &1.98
 \\
\texttt{llama-2-7b-chat}& 3.67 & 3.79 &4.89 &3.88 &3.28 &3.98
 \\ 
\texttt{llama-2-13b-chat} & 3.08 &3.48 &4.38 &3.78 &3.08 &3.48
 \\
\texttt{llama-2-70b-chat} & 2.68 &2.89 &3.88 &3.73 &2.98 &3.43
 \\ 
\hline \hline
\multicolumn{7}{|c|}{\textit{\cellcolor{yellow!10}Type 4: If you were to meet the CEO of the company, what topics would you like to discuss?}} \\ \hline \hline
\cellcolor{gray!20}Human & \cellcolor{gray!20}2.80 & \cellcolor{gray!20}3.12 & \cellcolor{gray!20}3.23 & \cellcolor{gray!20}3.45 & \cellcolor{gray!20}2.56 & \cellcolor{gray!20}2.80
\\ \hline \hline
\texttt{text-ada-001} &  3.89 &4.45 &4.67 &4.19 &4.24 &4.69
\\ 
\texttt{text-curie-001} & 3.64 &4.24 &4.14 &3.74 &3.24 &3.74
 \\
\texttt{text-babbage-001} &  3.77 &4.37 &4.19 &3.99 &3.59 &3.94
\\ 
\texttt{text-davinci-003} &  3.54 &4.19 &3.98 &2.98 &3.09 &3.69
\\
\texttt{GPT-3.5 Turbo} & 2.01 &2.67 &3.42 &3.94 &2.83 &2.11
 \\ 
\texttt{GPT-4 Turbo} & 2.07 &2.72 &3.27 &3.98 &2.47 &2.12
 \\
\texttt{llama-2-7b-chat}& 3.94 &4.05 &4.54 &4.58 &4.14 &3.98
 \\ 
\texttt{llama-2-13b-chat} & 3.54 &3.94 &4.14 &4.05 &3.01 &3.74
 \\
\texttt{llama-2-70b-chat} & 3.44 &3.74 &3.94 &3.94 &2.94 &3.34
\\ 
\hline
\end{tabular}
}
\caption{\label{tab_r1}
Comparison of assigned scores for speech transcripts across four question types between human evaluators and LLMs based on six human evaluation criteria.
}
\end{table*}

\subsection{Analyzing the performance of various LLMs in comparison to human evaluation in \textit{identifying instances of errors} (\textit{Task B}) and \textit{providing constructive feedback and practical suggestions} (\textit{Task C})} \label{app2}

In this section, we discuss the performance of various LLMs in comparison to human evaluation in \textit{\underline{identifying instances of errors}} (\textbf{\textit{Task B}}) and \textit{\underline{providing constructive feedback and practical} \underline{suggestions}} (\textbf{\textit{Task C}}).

\subsubsection{Type 1: \textit{\colorbox{blue!10}{Tell me about yourself.}}}
The evaluation is based on the spoken transcript (student's response) contained in the prompt shown in Figure \ref{fig6}. 
\\  \\
\textbf{Output Figures:}
\begin{itemize}

\item Figure \ref{fig7}: GPT-3.5 Turbo evaluation of the response to \textit{Question Type 1}.

\item Figure \ref{fig8}: GPT-4 Turbo evaluation of the response to \textit{Question Type 1}.

\item Figure \ref{fig9}: Human evaluation of the response to \textit{Question Type 1}.

\end{itemize}
\subsubsection{Type 2: \textit{\colorbox{green!10}{What are your strengths and} \colorbox{green!10}{weaknesses?}}}
The evaluation is based on the spoken transcript (student's response) contained in the prompt shown in Figure \ref{fig10}.
\\  \\
\textbf{Output Figures:}

\begin{itemize}
\item Figure \ref{fig11}: GPT-3.5 Turbo evaluation of the response to \textit{Question Type 2}.
\item Figure \ref{fig12}: GPT-4 Turbo evaluation of the response to \textit{Question Type 2}.
\item Figure \ref{fig22}: Human evaluation of the response to \textit{Question Type 2}.

\end{itemize}
\subsubsection{Type 3: \textit{\colorbox{red!10}{What are your hobbies and} \colorbox{red!10}{leisure interests?}}}
The evaluation is based on the spoken transcript (student's response) contained in the prompt shown in Figure \ref{fig13}.
\\ \\
\textbf{Output Figures:}

\begin{itemize}
\item Figure \ref{fig18}: GPT-3.5 Turbo evaluation of the response to \textit{Question Type 3}.
\item Figure \ref{fig19}: GPT-4 Turbo evaluation of the response to \textit{Question Type 3}.
\item Figure \ref{fig20}: Human evaluation of the response to \textit{Question Type 3}.
\end{itemize}
\subsubsection{Type 4: \textit{\colorbox{yellow!10}{If you were to meet the CEO of} \colorbox{yellow!10}{the company, what topics would you like} \colorbox{yellow!10}{to discuss?}}}
The evaluation is based on the spoken transcript (student's response) contained in the prompt shown in Figure \ref{fig14}.
\\ \\
\textbf{Output Figures:}

\begin{itemize}

\item Figure \ref{fig15}: GPT-3.5 Turbo evaluation of the response to \textit{Question Type 4}.
\item Figure \ref{fig16}: GPT-4 Turbo evaluation of the response to \textit{Question Type 4}.
\item Figure \ref{fig17}: Human evaluation of the response to \textit{Question Type 4}.

\end{itemize}

In a general analysis of the evaluations provided for various types of questions, it is evident that, while GPT-3.5 Turbo and GPT-4 Turbo offer valuable insights, the feedback they provide falls short in comparison to human evaluators. The discrepancies are noticeable across six human evaluation criteria, including \textit{fluency}, \textit{coherence}, \textit{tone/politeness}, \textit{relevance}, \textit{conciseness}, and \textit{grammaticality}.

\textbf{\textit{Fluency}:}
The GPT models (i.e., GPT-3.5 Turbo and GPT-4 Turbo) generally identify issues with fluency, such as awkward phrasing or run-on sentences. However, their feedback lacks specificity, often missing concrete examples that human evaluators easily identify. Humans excel in pinpointing instances of repetition, filler words, and disruptions in flow, showcasing a more refined understanding of spoken language nuances.

\textbf{\textit{Coherence}:}
GPT-3.5 Turbo and GPT-4 Turbo recognize general problems with coherence, but their feedback lacks precision. Human evaluators excel in highlighting specific instances of unclear transitions, disjointed phrases, and structural issues. This demonstrates that the contextual understanding required to deliver a coherent response is an area in which LLMs currently struggle.

\textbf{\textit{Tone/Politeness}:}
GPT-3.5 Turbo and GPT-4 Turbo demonstrate performance on par with humans. Although both LLMs generally offer generic feedback on tone/politeness, they may lack nuanced insights in some cases. Human evaluators excel at discerning informal or inappropriate language choices, providing specific examples, and suggesting improvements to maintain a professional tone. In general, GPT-4 Turbo performs almost on par with human evaluators, although GPT-3.5 Turbo lags in providing critical suggestions to improve the tone of the response.

\textbf{\textit{Relevance}:}
The GPT models (i.e., GPT-3.5 Turbo and GPT-4 Turbo) offer feedback on relevance, but their responses are often broad and lack specificity. Human evaluators excel in identifying precisely which details are off-topic or irrelevant, providing targeted advice on how to focus on the most pertinent information for a given context.

\textbf{\textit{Conciseness}:}
While GPT models (i.e., GPT-3.5 Turbo and GPT-4 Turbo) recognize verbosity, their feedback tends to be general. Human evaluators excel in pinpointing excessive details, redundancy, and unnecessary elaboration. They offer concrete examples and specific advice on how to streamline information while maintaining conciseness.

\textbf{\textit{Grammaticality}:}
GPT-3.5 Turbo and GPT-4 Turbo can identify some grammatical errors, but their feedback is often less precise. Human evaluators excel in providing specific examples of errors, ranging from incorrect verb forms to missing conjunctions, and offer detailed suggestions for improvement.

In conclusion, while GPT-3.5 Turbo and GPT-4 Turbo provide valuable automated evaluations, the depth and specificity of the feedback provided by human evaluators remain unmatched. Humans demonstrate a superior ability to understand context, identify all instances of errors, and offer tailored suggestions for improvement. The existing gaps between LLMs and human evaluators underscore the need for further advancements in natural language processing and additional progress in LLMs to achieve more nuanced and context-aware language understanding.


\begin{figure*}
    \centering
    \begin{tcolorbox}[enhanced,fit to height=24.5cm,
  colback=violet!20!white!10!white,colframe= violet!65!white,title= Prompt to evaluate a response to \textit{Question Type 1} across six human evaluation criteria, drop fuzzy shadow]

  {\fontfamily{qcr}\selectfont
 
Imagine you are a mentor assessing a student's response in an HR interview. Your job is to evaluate their response, indicating errors, and providing comprehensive feedback within a 0-5 score range across various criteria. \\

Interview Question:

\{Tell me about yourself.\}\\

Student's Response:

\{Thank you for a long night. Introduce myself myself. Priya Sahu are belong to Manda, Madhya Pradesh.Now talking about my qualification. Currently I am pursuing I have to completed my graduation battles of technology into Department of Petrochemical Engineering from University Institute of Technology RGPV Bhopal. Apart from this I have completed my in plant training at Rashtriya Chemicals and Fertilizers Limited RCF alibag. Along with this I have completed my internship from our doubt nut Azar doubt solving expertsinarmas within a month of a 2 period I have completed my 10th and.Well from Bharati High secondary school Malad.Image of Temple 82 and 12th age 86\%.Now you are moving towards my.Extracurricular activities I love singing and a world painting even I have go through the Indian Idol contestant and now.\}\\

Assessment Criteria: \\

Fluency (To what extent does the candidate articulate responses smoothly, ranging from interrupted expression to exceptionally smooth articulation?):

1. Rate the candidate's fluency on a scale of 0 to 5 (0 being disfluent, 5 being exceptionally fluent).

2. Identify all instances of disfluency.

3. Offer suggestions for improving overall fluency and flow.\\

Coherence (To what extent is the
candidate’s response transparent and logically
structured?):

1. Rate the candidate's coherence on a scale of 0 to 5 (0 being incoherent, 5 being highly coherent).

2. Identify all instances where the response lacks coherence.

3. Provide suggestions for improving coherence and maintaining logical flow.\\

Tone/Politeness (How does the candidate’s language exhibit the level of formality, respect, and professionalism suitable for an HR interview setting?):

1. Rate the candidate's tone and politeness on a scale of 0 to 5 (0 representing impolite or unprofessional language, 5 denoting high politeness and professionalism).

2. Identify all instances where the response lacks politeness or professionalism.

3. Offer suggestions for maintaining a respectful and professional tone.\\

Relevance (How well does the candidate’s response directly address and align with the given interview question?):

1. Rate the candidate's relevance to the interview question on a scale of 0 to 5 (0 representing a lack of relevance, 5 indicating a highly relevant response).

2. Identify all off-topic or irrelevant elements in the response.

3. Offer suggestions for maintaining focus and relevance in interview responses.\\

Conciseness (How effectively does the candidate deliver information in a brief, yet informative manner, avoiding unnecessary verbosity?):

1. Rate the candidate's conciseness on a scale of 0 to 5 (0 being overly verbose, 5 being appropriately concise).

2. Identify all instances where the response is excessively wordy or lacks brevity.

3. Offer suggestions for delivering information in a more concise manner.\\

 Grammaticality (How grammatically correct is the language used in the candidate’s response?):

1. Rate the candidate's grammatical accuracy on a scale of 0 to 5 (0 indicating numerous grammatical errors, 5 indicating impeccable grammar).

2. Identify all grammatical errors in the response.

3. Offer suggestions for grammatical improvements.\\

[NOTE: Please provide the evaluation report in the same format as above.]

}
  
\end{tcolorbox}

    \caption{Prompt to evaluate a response to \textit{Question Type 1} across six human evaluation criteria.}
    \label{fig6}
\end{figure*}


\begin{figure*}
    \centering
    \begin{tcolorbox}[enhanced,fit to height=20cm,
  colback=blue!20!white!10!white,colframe= blue!60!white,title= GPT-3.5 Turbo evaluation of the response to \textit{Question Type 1}, drop fuzzy shadow]

  {\fontfamily{qcr}\selectfont
 
Assessment Criteria:\\

Fluency:

1. Rating: 3

2. Instances of disfluency: The response is generally fluent but has some awkward phrasing, affecting overall flow.

3. Improvement suggestions: Practice more fluid transitions between ideas. Avoid unnecessary pauses and work on delivering responses with more natural flow. \\

Coherence:

1. Rating: 3

2. Instances where the response lacks coherence: The response lacks coherence in certain sentences, affecting overall coherence. For example, "Within a month of a 2 period" is unclear.

3. Improvement suggestions: Focus on expressing ideas clearly and logically. Clarify ambiguous phrases for a more coherent narrative.
\\

Tone/Politeness:

1. Rating: 4

2. Instances where the response lacks politeness or professionalism: The tone is generally polite and professional.

3. Improvement suggestions:  Maintain this level of politeness throughout. No major issues identified in tone or politeness.
\\

Relevance:

1. Rating: 3

2. Instances of off-topic or irrelevant elements: The response contains relevant information about education and extracurricular activities. However, there are some unclear phrases affecting relevance.

3. Improvement suggestions: Remove unclear phrases for better relevance. Provide more details about specific achievements in education and extracurricular activities.
\\

Conciseness:

1. Rating: 2

2. Instances of excessive verbosity: The response is somewhat verbose and could be more concise. The sentence structure could be improved for brevity.

3. Improvement suggestions: Streamline information for brevity while maintaining completeness. Use concise language to convey the same information more efficiently.
\\

 Grammaticality:

1. Rating: 3

2. Instances of grammatical errors: There are grammatical errors, such as "Within a month of a 2 period," which is unclear.

3. Improvement suggestions: Review grammar rules, focus on sentence structure, and clarify expressions for better understanding.\\

}
  
\end{tcolorbox}

    \caption{GPT-3.5 Turbo evaluation of the response to \textit{Question Type 1}.}
    \label{fig7}
\end{figure*}


\begin{figure*}
    \centering
    \begin{tcolorbox}[enhanced,fit to height=21cm,
  colback=blue!20!white!10!white,colframe= blue!60!white,title= GPT-4 Turbo evaluation of the response to \textit{Question Type 1}, drop fuzzy shadow]

  {\fontfamily{qcr}\selectfont
 
Assessment Criteria:\\

Fluency:

1. Rating: 2

2. Instances of disfluency: The response has several instances of incoherence, making it challenging to follow the candidate's narrative smoothly. The phrase "I have to completed my graduation battles of technology into Department of Petrochemical Engineering" lacks coherence and disrupts the flow.

3. Improvement suggestions: Practice speaking more slowly and enunciating words for better coherence. Organize thoughts before speaking to avoid interruptions and stammering. \\

Coherence:

1. Rating: 2

2. Instances where the response lacks coherence: The candidate's transition from talking about graduation to in-plant training is abrupt and unclear. The mention of "our doubt nut Azar doubt solving expertsinarmas" is confusing and lacks context.

3. Improvement suggestions: Create a clear and logical structure for the response, ensuring smooth transitions between different topics. Avoid using unclear or jargon-like phrases; strive for simplicity and coherence.\\

Tone/Politeness:

1. Rating: 3

2. Instances where the response lacks politeness or professionalism: The phrase "Thank you for a long night" may be interpreted as informal and might not be the best way to start a professional interview.

3. Improvement suggestions: Begin with a more standard and professional greeting, expressing gratitude for the opportunity to interview.\\

Relevance:

1. Rating: 2

2. Instances of off-topic or irrelevant elements: The mention of the candidate's percentage in 10th and 12th grades, as well as the image of a temple, is not directly relevant to the question.

3. Improvement suggestions:  Focus on providing information directly related to professional background, skills, and experiences. Avoid including personal details that do not contribute to answering the question.\\

Conciseness:

1. Rating: 1

2. Instances of excessive verbosity: The response is overly detailed and includes information that may not be crucial for the initial "Tell me about yourself" question.

3. Improvement suggestions: Prioritize key details and accomplishments, keeping the response concise and relevant to the question.\\

 Grammaticality:

1. Rating: 2

2. Instances of grammatical errors: "I have to completed my graduation battles of technology into Department of Petrochemical Engineering" contains grammatical errors. Several sentences lack proper punctuation and could benefit from restructuring.

3. Improvement suggestions: Review basic grammar rules, particularly sentence structure and punctuation.\\

}
  
\end{tcolorbox}

    \caption{GPT-4 Turbo evaluation of the response to \textit{Question Type 1}.}
    \label{fig8}
\end{figure*}


\begin{figure*}
    \centering
    \begin{tcolorbox}[enhanced,fit to height=24cm,
  colback=green!20!white!10!white,colframe= blue!60!white,title= Human evaluation of the response to \textit{Question Type 1}, drop fuzzy shadow]

  {\fontfamily{qcr}\selectfont
 
Assessment Criteria:\\

Fluency:

1. Rating: 1

2. Instances of disfluency: The response contains several instances of repetition (e.g., "Introduce myself myself"), incorrect verb forms (e.g., "are belong to"), and disjointed phrases (e.g., "Currently I am pursuing I have to completed my graduation battles of technology into Department of Petrochemical Engineering"). These issues disrupt the flow of the response.

3. Improvement suggestions: To improve fluency, the candidate should practice delivering their introduction in a smooth, uninterrupted manner. They could structure their thoughts beforehand and focus on using correct verb tenses and avoiding repetition. A rephrased sentence might be: "My name is Priya Sahu, and I'm from Manda, Madhya Pradesh. I am currently completing my Bachelor of Technology in Petrochemical Engineering at the University Institute of Technology RGPV in Bhopal."\\

Coherence:

1. Rating: 1

2. Instances where the response lacks coherence: The response contains an unclear transition between education and extracurricular activities, as evidenced by the statement "Now you are moving towards my." Additionally, the information provided appears muddled, as seen in phrases like "battles of technology into Department" and "from our doubt nut Azar doubt solving expertsinarmas within a month of a 2 period."

3. Improvement suggestions: Use clear transitions between topics. Ensure that each sentence adds to the narrative in a logical sequence. Consider rephrasing it to "I have pursued my Bachelor's of Technology with a focus on petrochemical engineering, complemented by hands-on training and internships in the industry. Beyond academics, I have a passion for the arts, particularly singing and painting, which I have showcased in competitions such as Indian Idol." \\

Tone/Politeness:

1. Rating: 2

2. Instances where the response lacks politeness or professionalism: The opening "Thank you for a long night" could be interpreted as unprofessional or sarcastic. The overall tone is casual and lacks the formality expected in an interview.

3. Improvement suggestions: The candidate should start with a more conventional greeting, such as "Thank you for this opportunity to introduce myself." This sets a professional tone from the beginning.\\

Relevance:

1. Rating: 2

2. Instances of off-topic or irrelevant elements: The mention of Indian Idol is not directly relevant to the professional introduction unless it directly relates to skills or experiences applicable to the job (e.g.,"I love singing and a world painting even I have go through the Indian Idol contestant.").

3. Improvement suggestions: The candidate should focus on experiences and qualifications that are directly relevant to the job they are applying for. For example, they could omit the Indian Idol reference unless it demonstrates qualities such as perseverance or public speaking skills that are relevant to the position. \\

Conciseness:

1. Rating: 1

2. Instances of excessive verbosity: The response includes unnecessary details and convoluted explanations (e.g., the detailed percentages from high school "Image of Temple 82 and 12th age 86\%.", "in plant training at Rashtriya Chemicals and Fertilizers Limited RCF alibag").

3. Improvement suggestions: Stick to the most impactful and relevant information. Briefly mention personal interests or achievements only if they demonstrate qualities valuable to the position. The candidate should aim to be more succinct by focusing on key points and eliminating extraneous information. For instance, "I completed in-plant training at Rashtriya Chemicals and Fertilizers Limited in Alibag.\\

 Grammaticality:

1. Rating: 1

2. Instances of grammatical errors: The response contains numerous grammatical mistakes, such as incorrect verb conjugation (e.g.,"Priya Sahu are belong to Manda") and incorrect past tense usage (e.g., "I have to completed my graduation").

3. Improvement suggestions: The candidate should review basic grammar rules and practice constructing sentences correctly. Consider rephrasing it to "I belong to Manda, Madhya Pradesh," and "I have completed my graduation in Petrochemical Engineering."\\

}
  
\end{tcolorbox}

    \caption{Human evaluation of the response to \textit{Question Type 1}.}
    \label{fig9}
\end{figure*}



\begin{figure*}
    \centering
    \begin{tcolorbox}[enhanced,fit to height=24cm,
  colback=violet!20!white!10!white,colframe= violet!65!white,title= Prompt to evaluate a response to \textit{Question Type 2} across six human evaluation criteria, drop fuzzy shadow]

  {\fontfamily{qcr}\selectfont
 
Imagine you are a mentor assessing a student's response in an HR interview. Your job is to evaluate their response, indicating errors, and providing comprehensive feedback within a 0-5 score range across various criteria. \\

Interview Question:

\{What are your strengths and weaknesses?\}\\

Student's Response:

\{So talking about my strength, my biggest strength is a multitasking person because I have worked in a situation like where I have to deal with different different things at the same time, like for an example, I was digital marketing also and intern in event managing also. So at the same time I manage both the things and at that time I was in college so doing assignments and.Attending exams over also the task for me, but I manage all things with proper timings and.Never having excuses for the same and as of my I'm not having any kind of weakness. But one thing is that and it perfection in everything. So this perfection sometimes consumes a lot of time, so there is time limit also. So I need to take care of that.\}\\

Assessment Criteria: \\

Fluency (To what extent does the candidate articulate responses smoothly, ranging from interrupted expression to exceptionally smooth articulation?):

1. Rate the candidate's fluency on a scale of 0 to 5 (0 being disfluent, 5 being exceptionally fluent).

2. Identify all instances of disfluency.

3. Offer suggestions for improving overall fluency and flow.\\

Coherence (To what extent is the
candidate’s response transparent and logically
structured?):

1. Rate the candidate's coherence on a scale of 0 to 5 (0 being incoherent, 5 being highly coherent).

2. Identify all instances where the response lacks coherence.

3. Provide suggestions for improving coherence and maintaining logical flow.\\

Tone/Politeness (How does the candidate’s language exhibit the level of formality, respect, and professionalism suitable for an HR interview setting?):

1. Rate the candidate's tone and politeness on a scale of 0 to 5 (0 representing impolite or unprofessional language, 5 denoting high politeness and professionalism).

2. Identify all instances where the response lacks politeness or professionalism.

3. Offer suggestions for maintaining a respectful and professional tone.\\

Relevance (How well does the candidate’s response directly address and align with the given interview question?):

1. Rate the candidate's relevance to the interview question on a scale of 0 to 5 (0 representing a lack of relevance, 5 indicating a highly relevant response).

2. Identify all off-topic or irrelevant elements in the response.

3. Offer suggestions for maintaining focus and relevance in interview responses.\\

Conciseness (How effectively does the candidate deliver information in a brief, yet informative manner, avoiding unnecessary verbosity?):

1. Rate the candidate's conciseness on a scale of 0 to 5 (0 being overly verbose, 5 being appropriately concise).

2. Identify all instances where the response is excessively wordy or lacks brevity.

3. Offer suggestions for delivering information in a more concise manner.\\

 Grammaticality (How grammatically correct is the language used in the candidate’s response?):

1. Rate the candidate's grammatical accuracy on a scale of 0 to 5 (0 indicating numerous grammatical errors, 5 indicating impeccable grammar).

2. Identify all grammatical errors in the response.

3. Offer suggestions for grammatical improvements.\\

[NOTE: Please provide the evaluation report in the same format as above.]

}
  
\end{tcolorbox}

    \caption{Prompt to evaluate a response to \textit{Question Type 2} across six human evaluation criteria.}
    \label{fig10}
\end{figure*}


\begin{figure*}
    \centering
    \begin{tcolorbox}[enhanced,fit to height=18cm,
  colback=blue!20!white!10!white,colframe= blue!60!white,title= GPT-3.5 Turbo evaluation of the response to \textit{Question Type 2}, drop fuzzy shadow]

  {\fontfamily{qcr}\selectfont
 
Assessment Criteria:\\

Fluency:

1. Rating: 3

2. Instances of disfluency: The response shows moderate fluency, but there are instances of run-on sentences that can affect coherence.

3. Improvement suggestions: Break down complex sentences into shorter, clearer ones to enhance overall fluency.\\

Coherence:

1. Rating: 3

2. Instances where the response lacks coherence: The response lacks coherence in some parts due to the run-on sentences and minor grammar issues.

3. Improvement suggestions: Focus on structuring sentences more coherently. Ensure that each point is expressed clearly.\\

Tone/Politeness:

1. Rating: 4

2. Instances where the response lacks politeness or professionalism: The tone is generally polite and professional.

3. Improvement suggestions: Sustain the positive tone and consider refining expressions for added professionalism.\\

Relevance:

1. Rating: 4

2. Instances of off-topic or irrelevant elements: The response is mostly relevant, but some details could be more focused.

3. Improvement suggestions: Trim unnecessary details and emphasize key points for a more impactful response.\\

Conciseness:

1. Rating: 3

2. Instances of excessive verbosity: The response tends to be lengthy, affecting conciseness.

3. Improvement suggestions: Streamline information and remove redundant details to create a more concise and focused response.\\

 Grammaticality:

1. Rating: 3

2. Instances of grammatical errors: There are grammatical errors and issues with sentence structure.

3. Improvement suggestions: Pay attention to grammar, especially sentence structure. Review and revise for coherence.\\

}
  
\end{tcolorbox}

    \caption{GPT-3.5 Turbo evaluation of the response to \textit{Question Type 2}.}
    \label{fig11}
\end{figure*}


\begin{figure*}
    \centering
    \begin{tcolorbox}[enhanced,fit to height=21cm,
  colback=blue!20!white!10!white,colframe= blue!60!white,title= GPT-4 Turbo evaluation of the response to \textit{Question Type 2}, drop fuzzy shadow]

  {\fontfamily{qcr}\selectfont
 
Assessment Criteria:\\

Fluency:

1. Rating: 2

2. Instances of disfluency: The response contains run-on sentences and lacks punctuation, making it difficult to follow. The use of "like" as a filler word and the phrase "digital marketing also and intern in event managing also" disrupt the flow.

3. Improvement suggestions: Practice speaking in complete sentences and use pauses to enhance understanding. Avoid filler words and ensure that each point is clearly separated from the next.\\

Coherence:

1. Rating: 2

2. Instances where the response lacks coherence: The response is jumbled, with the mention of multitasking, digital marketing, event management, and college responsibilities all in one breath. The weakness is not clearly stated, and the explanation is convoluted.

3. Improvement suggestions: Clearly separate strengths from weaknesses and provide a structured response. Use examples to illustrate points and ensure that the weakness is clearly identified and explained.\\

Tone/Politeness:

1. Rating: 4

2. Instances where the response lacks politeness or professionalism: The overall tone is casual but not impolite. The phrase "I'm not having any kind of weakness" could be perceived as lacking humility.

3. Improvement suggestions: Maintain a professional tone by acknowledging areas for improvement and avoid absolute statements that suggest infallibility.\\

Relevance:

1. Rating: 3

2. Instances of off-topic or irrelevant elements: The response includes relevant examples of multitasking but veers off-topic with the lengthy explanation of managing college responsibilities.

3. Improvement suggestions: Focus on providing examples directly related to professional strengths and weaknesses. Keep personal academic experiences brief unless they directly relate to job performance.\\

Conciseness:

1. Rating: 2

2. Instances of excessive verbosity: The response is verbose, with unnecessary repetition and lack of clear structure.

3. Improvement suggestions: Be more succinct by stating one strength and one weakness, each followed by a brief example or explanation.\\

 Grammaticality:

1. Rating: 2

2. Instances of grammatical errors:  Incorrect verb tense ("I was digital marketing also"), missing conjunctions ("and intern in event managing also"), and missing articles ("the task for me").

3. Improvement suggestions: Review basic grammar rules and practice constructing sentences that are grammatically correct. Consider writing out responses to practice proper structure.\\

}
  
\end{tcolorbox}

    \caption{GPT-4 Turbo evaluation of the response to \textit{Question Type 2}.}
    \label{fig12}
\end{figure*}


\begin{figure*}
    \centering
    \begin{tcolorbox}[enhanced,fit to height=24cm,
  colback=green!20!white!10!white,colframe= blue!60!white,title= Human evaluation of the response to \textit{Question Type 2}, drop fuzzy shadow]

  {\fontfamily{qcr}\selectfont
 
Assessment Criteria:\\

Fluency:

1. Rating: 2

2. Instances of disfluency: The response includes filler words (e.g., "like", "so") and repetition (e.g., "different different things", "also and intern", "and.Attending exams over also the task for me").  The response also contains several run-on sentences and lacks pauses, which disrupts the flow (e.g., "like for an example, I was digital marketing also and intern in event managing also.").

3. Improvement suggestions: To improve fluency, the student should practice speaking in complete sentences and use pauses appropriately. Eliminating filler words and repetition will also help. Speaking slowly and deliberately can aid in maintaining a steady flow. Consider rephrasing it to, "For instance, I have experience in digital marketing and event management, where I successfully juggled multiple tasks." \\

Coherence:

1. Rating: 2

2. Instances where the response lacks coherence: The response is somewhat disorganized, with the student jumping from one point to another (e.g., "I was digital marketing also and intern in event managing also"). The use of the phrase "as of my" is unclear and the connection between perfectionism and time management is not well explained.

3. Improvement suggestions: The student should structure the response by first discussing strengths in a clear manner, then moving on to weaknesses. They should also clarify the statements and ensure that each point follows logically from the previous one. Consider rephrasing as "My primary strength is multitasking. I have managed simultaneous projects in digital marketing and event management while attending college. Regarding weaknesses, I tend to seek perfection, which can be time-consuming."\\

Tone/Politeness:

1. Rating: 3

2. Instances where the response lacks politeness or professionalism: There is a minor issue with the statement "I'm not having any kind of weakness," which could come across as lacking self-awareness or humility.

3. Improvement suggestions: The student should maintain a humble tone, acknowledging that everyone has weaknesses and focusing on how they are addressing them. Rephrase to "While I strive to perform well in all areas, I am aware that there is always room for growth."\\

Relevance:

1. Rating: 3

2. Instances of off-topic or irrelevant elements: The response is mostly relevant, but the student's example about managing work and studies does not directly address the question of strengths and weaknesses.

3. Improvement suggestions: The student should directly state their strengths and weaknesses related to the job they are applying for and avoid lengthy examples that do not add to the answer's relevance. Consider rephrasing it as "In addition to my multitasking ability, I am also detail-oriented, which is a strength in managing complex projects. However, my attention to detail can sometimes slow me down and I am working to find the right balance."\\

Conciseness:

1. Rating: 2

2. Instances of excessive verbosity: The response includes unnecessary repetition (e.g., "like for an example", "at that time I was in college so doing assignments and.Attending exams over also the task for me") and could be more concise (e.g., "like where I have to deal with different different things at the same time" could be shortened to "where I managed multiple tasks simultaneously.").

3. Improvement suggestions: The student should practice delivering their response in a more straightforward manner, avoiding repetition, and focusing on key points. Consider rephrasing it to "My strength is multitasking, as demonstrated by balancing digital marketing, event management, and college studies. My pursuit of perfection can lead to time constraints, which I am learning to manage more effectively."\\

 Grammaticality:

1. Rating: 2

2. Instances of grammatical errors: The response contains several grammatical mistakes (e.g., "my biggest strength is a multitasking person", incorrect use of "like", "I was digital marketing also and intern", "Attending exams over also the task for me", "as of my I'm not having any kind of weakness", "and it perfection in everything").

3. Improvement suggestions: The student should review basic grammar rules and practice constructing sentences correctly. They may benefit from speaking or writing practice exercises and using language learning tools. Consider rephrasing it: "I have experience in digital marketing and have also interned in event management. Additionally, I managed my college assignments and exams." \\

}
  
\end{tcolorbox}

    \caption{Human evaluation of the response to \textit{Question Type 2}.}
    \label{fig22}
\end{figure*}



\begin{figure*}
    \centering
    \begin{tcolorbox}[enhanced,fit to height=24cm,
  colback=violet!20!white!10!white,colframe= violet!65!white,title= Prompt to evaluate a response to \textit{Question Type 3} across six human evaluation criteria, drop fuzzy shadow]

  {\fontfamily{qcr}\selectfont
 
Imagine you are a mentor assessing a student's response in an HR interview. Your job is to evaluate their response, indicating errors, and providing comprehensive feedback within a 0-5 score range across various criteria. \\

Interview Question:

\{What are your hobbies and leisure interests?\}\\

Student's Response:

\{Basically my hobbies, I just wanted to solve different kind of problems. I'm very much into puzzles and all, and along with my technical side realise a small side of my creativity. While I love to enjoy drawing and I love to love painting in sketching in all and all. After that I also love cooking. I am very much into these type of things and interest. I am very much interested in all the space things space objects such as I I just love.Quantum physics I love to study about the black hole and new creatures that we are going. We are just. I just love to study about the space things just like we are exploring the space, exploring the universe. How many universe are there in? How many galaxies are there?\}\\

Assessment Criteria: \\

Fluency (To what extent does the candidate articulate responses smoothly, ranging from interrupted expression to exceptionally smooth articulation?):

1. Rate the candidate's fluency on a scale of 0 to 5 (0 being disfluent, 5 being exceptionally fluent).

2. Identify all instances of disfluency.

3. Offer suggestions for improving overall fluency and flow.\\

Coherence (To what extent is the
candidate’s response transparent and logically
structured?):

1. Rate the candidate's coherence on a scale of 0 to 5 (0 being incoherent, 5 being highly coherent).

2. Identify all instances where the response lacks coherence.

3. Provide suggestions for improving coherence and maintaining logical flow.\\

Tone/Politeness (How does the candidate’s language exhibit the level of formality, respect, and professionalism suitable for an HR interview setting?):

1. Rate the candidate's tone and politeness on a scale of 0 to 5 (0 representing impolite or unprofessional language, 5 denoting high politeness and professionalism).

2. Identify all instances where the response lacks politeness or professionalism.

3. Offer suggestions for maintaining a respectful and professional tone.\\

Relevance (How well does the candidate’s response directly address and align with the given interview question?):

1. Rate the candidate's relevance to the interview question on a scale of 0 to 5 (0 representing a lack of relevance, 5 indicating a highly relevant response).

2. Identify all off-topic or irrelevant elements in the response.

3. Offer suggestions for maintaining focus and relevance in interview responses.\\

Conciseness (How effectively does the candidate deliver information in a brief, yet informative manner, avoiding unnecessary verbosity?):

1. Rate the candidate's conciseness on a scale of 0 to 5 (0 being overly verbose, 5 being appropriately concise).

2. Identify all instances where the response is excessively wordy or lacks brevity.

3. Offer suggestions for delivering information in a more concise manner.\\

 Grammaticality (How grammatically correct is the language used in the candidate’s response?):

1. Rate the candidate's grammatical accuracy on a scale of 0 to 5 (0 indicating numerous grammatical errors, 5 indicating impeccable grammar).

2. Identify all grammatical errors in the response.

3. Offer suggestions for grammatical improvements.\\

[NOTE: Please provide the evaluation report in the same format as above.]

}
  
\end{tcolorbox}

    \caption{Prompt to evaluate a response to \textit{Question Type 3} across six human evaluation criteria.}
    \label{fig13}
\end{figure*}


\begin{figure*}
    \centering
    \begin{tcolorbox}[enhanced,fit to height=18cm,
  colback=blue!20!white!10!white,colframe= blue!60!white,title= GPT-3.5 Turbo evaluation of the response to \textit{Question Type 3}, drop fuzzy shadow]

  {\fontfamily{qcr}\selectfont
 
Assessment Criteria:\\

Fluency:

1. Rating: 3

2. Instances of disfluency: The response has some interruptions and instances of incoherence, affecting overall fluency.

3. Improvement suggestions: The candidate should work on smoother transitions between ideas, reducing pauses, and practicing to enhance overall fluency.\\

Coherence:

1. Rating: 2

2. Instances where the response lacks coherence: The response lacks a clear structure, and the ideas are presented in a somewhat disjointed manner.

3. Improvement suggestions: The candidate should organize their thoughts more logically. Breaking down the response into distinct sections, such as puzzles, artistic interests, and space exploration, would improve coherence.\\

Tone/Politeness:

1. Rating: 4

2. Instances where the response lacks politeness or professionalism: The tone is generally polite; however, there is room for improvement in terms of formality to enhance professionalism.

3. Improvement suggestions: Maintaining a consistently formal tone throughout and expressing enthusiasm in a more structured manner would elevate the professionalism of the response.\\

Relevance:

1. Rating: 3

2. Instances of off-topic or irrelevant elements: The response contains relevant information, but it becomes somewhat scattered due to the lack of a clear focus on specific hobbies.

3. Improvement suggestions: The candidate should prioritize and elaborate on a couple of main hobbies to provide a more focused and detailed response.\\

Conciseness:

1. Rating: 2

2. Instances of excessive verbosity: The response is somewhat verbose, with repetitive phrases and excessive details.

3. Improvement suggestions: The candidate should aim for brevity by highlighting key points and avoiding unnecessary repetition. Providing concise details about each hobby would enhance the overall response.\\

 Grammaticality:

1. Rating: 3

2. Instances of grammatical errors: The response has several grammatical errors and awkward phrasings.

3. Improvement suggestions: The candidate should review sentence structure and grammar. Proofreading and refining sentence construction will contribute to a more polished response.\\

}
  
\end{tcolorbox}

    \caption{GPT-3.5 Turbo evaluation of the response to \textit{Question Type 3}.}
    \label{fig18}
\end{figure*}


\begin{figure*}
    \centering
    \begin{tcolorbox}[enhanced,fit to height=18cm,
  colback=blue!20!white!10!white,colframe= blue!60!white,title=GPT-4 Turbo evaluation of the response to \textit{Question Type 3}, drop fuzzy shadow]

  {\fontfamily{qcr}\selectfont
 
Assessment Criteria:\\

Fluency:

1. Rating: 3

2. Instances of disfluency: The repetition of phrases like "I just love" and abrupt transitions, such as "I am very much into these type of things and interest," contributes to incoherence.

3. Improvement suggestions: Practice smoother transitions between ideas, and vary sentence structures to improve overall flow.\\

Coherence:

1. Rating: 2

2. Instances where the response lacks coherence: Abrupt shifts between discussing hobbies and space interests make it challenging to follow a logical flow. For instance, the sudden transition from drawing to "space things" lacks coherence.

3. Improvement suggestions: Organize the response with a clear structure, providing distinct sections for hobbies and space interests. Maintain a logical progression of ideas.\\

Tone/Politeness:

1. Rating: 4

2. Instances where the response lacks politeness or professionalism:  The response is generally polite but lacks a bit of professionalism due to the informal and somewhat chaotic delivery.

3. Improvement suggestions: Maintain a polite tone but strive for a more professional delivery by organizing your thoughts and speaking clearly.\\

Relevance:

1. Rating: 3

2. Instances of off-topic or irrelevant elements: The detailed discussion about space and quantum physics, while interesting, strays from the main topic of hobbies and leisure interests.

3. Improvement suggestions: Provide more depth on non-technical hobbies to offer a well-rounded response. Ensure that the response aligns with the interviewer's question.\\

Conciseness:

1. Rating: 2

2. Instances of excessive verbosity: The response is repetitive and includes unnecessary details, making it longer than needed.

3. Improvement suggestions: Be more succinct. Mention each hobby once and avoid going into too much detail unless prompted by the interviewer.\\

 Grammaticality:

1. Rating: 3

2. Instances of grammatical errors: There are grammatical errors, such as "I just love" (lacking space) and inconsistencies in tense usage, like "I AM very much into these type of things."

3. Improvement suggestions: Thorough proofreading is needed to correct errors. Maintain consistent tense usage and consider varying sentence structures for grammatical precision.\\

}
  
\end{tcolorbox}

    \caption{GPT-4 Turbo evaluation of the response to \textit{Question Type 3}.}
    \label{fig19}
\end{figure*}


\begin{figure*}
    \centering
    \begin{tcolorbox}[enhanced,fit to height=24cm,
  colback=green!20!white!10!white,colframe= blue!60!white,title= Human evaluation of the response to \textit{Question Type 3}, drop fuzzy shadow]

  {\fontfamily{qcr}\selectfont
 
Assessment Criteria:\\

Fluency:

1. Rating: 2

2. Instances of disfluency: The response includes several instances of repetition (e.g., "I love to love," "I am very much into these type of things and interest," "I just love.Quantum physics I love"), hesitation (e.g., "such as I I just love"), filler words (e.g., "and all and all"), and broken sentences (e.g., "How many universe are there in? How many galaxies are there?").

3. Improvement suggestions: To improve fluency, the candidate should practice speaking about their hobbies in a more structured manner. They could outline their main interests beforehand and avoid repetition by using synonyms or related terms. Additionally, slowing down the pace of speech can help reduce the use of fillers and hesitations. Use pauses instead of filler words to collect your thoughts.\\

Coherence:

1. Rating: 2

2. Instances where the response lacks coherence: The response jumps from one hobby to another without clear transitions (e.g., "While I love to enjoy drawing and I love to love painting in sketching in all and all. After that I also love cooking."). The sentence structure is also confusing at times, making it hard to follow (e.g., "How many universe are there in? How many galaxies are there?").

3. Improvement suggestions: The candidate should work on creating a more logical flow by connecting ideas with transitions. Consider rephrasing it to "In addition to my passion for puzzles, I also enjoy creative pursuits such as drawing, painting, and sketching. Furthermore, cooking allows me to express my creativity in the kitchen. My interest in space and quantum physics satisfies my curiosity about the universe."\\

Tone/Politeness:

1. Rating: 3

2. Instances where the response lacks politeness or professionalism: Excessive use of phrases like "I just love" may slightly diminish professionalism. For example, "I just love to study" might be refined. The response is generally polite but lacks professionalism due to the informal and somewhat chaotic nature of the delivery.

3. Improvement suggestions: Vary vocabulary to avoid repetition. Maintain enthusiasm while striking a balance with a more polished and professional tone. To maintain a professional tone, the candidate should aim for a more structured and formal response. They should avoid colloquial phrases and ensure that their enthusiasm does not detract from the professionalism of their delivery.\\

Relevance:

1. Rating: 3

2. Instances of off-topic or irrelevant elements:  The response veers off-topic when discussing quantum physics and space exploration, which, while interesting, are not directly related to hobbies or leisure interests unless the candidate actively engages in related activities in their free time.

3. Improvement suggestions: The candidate should focus on hobbies that they actively participate in and describe how these hobbies contribute to their personal development or relaxation. If they wish to mention interests like space, they should clarify how these interests manifest as hobbies (e.g., reading about space, attending lectures, etc.).\\

Conciseness:

1. Rating: 2

2. Instances of excessive verbosity: The response is verbose, with phrases like "I just love" being overly used. The response includes unnecessary repetition and filler words that make it wordy (e.g., "I am very much into these type of things and interest," "We are just," "I just love to study about the space things just like we are exploring the space, exploring the universe.").

3. Improvement suggestions: The candidate should aim to be more succinct by eliminating repetition and focusing on delivering each point clearly and directly. They could benefit from practicing concise responses to common interview questions.\\

 Grammaticality:

1. Rating: 2

2. Instances of grammatical errors: The response contains several grammatical errors, including incorrect verb tense (e.g., "realise" should be "realizing"), missing conjunctions (e.g., "I love to love painting in sketching in all and all"), and incorrect sentence structure (e.g., "How many universe are there in?").

3. Improvement suggestions: The candidate should review basic grammar rules and practice constructing sentences correctly. Reading their response aloud or writing it down could help identify and correct grammatical mistakes. Consider rephrasing it to  "I enjoy solving different kinds of problems, such as puzzles, which also allows me to express a bit of my creativity through drawing and painting. Additionally, I like cooking and have a keen interest in space exploration, including studying quantum physics and black holes." \\

}
  
\end{tcolorbox}

    \caption{Human evaluation of the response to \textit{Question Type 3}.}
    \label{fig20}
\end{figure*}



\begin{figure*}
    \centering
    \begin{tcolorbox}[enhanced,fit to height=24cm,
  colback=violet!20!white!10!white,colframe= violet!65!white,title= Prompt to evaluate a response to \textit{Question Type 4} across six human evaluation criteria, drop fuzzy shadow]

  {\fontfamily{qcr}\selectfont
 
Imagine you are a mentor assessing a student's response in an HR interview. Your job is to evaluate their response, indicating errors, and providing comprehensive feedback within a 0-5 score range across various criteria. \\

Interview Question:

\{If you were to meet the CEO of the company, what topics would you like to discuss?\}\\

Student's Response:

\{If I met CEO of the company, I would like to discuss about his vision about the future of the company.And I will discuss about my role in this vision, and I will tell him about the what I think about the future of the company. And lastly, I will try to get the feedback about my performance.That's it.\}\\

Assessment Criteria: \\

Fluency (To what extent does the candidate articulate responses smoothly, ranging from interrupted expression to exceptionally smooth articulation?):

1. Rate the candidate's fluency on a scale of 0 to 5 (0 being disfluent, 5 being exceptionally fluent).

2. Identify all instances of disfluency.

3. Offer suggestions for improving overall fluency and flow.\\

Coherence (To what extent is the
candidate’s response transparent and logically
structured?):

1. Rate the candidate's coherence on a scale of 0 to 5 (0 being incoherent, 5 being highly coherent).

2. Identify all instances where the response lacks coherence.

3. Provide suggestions for improving coherence and maintaining logical flow.\\

Tone/Politeness (How does the candidate’s language exhibit the level of formality, respect, and professionalism suitable for an HR interview setting?):

1. Rate the candidate's tone and politeness on a scale of 0 to 5 (0 representing impolite or unprofessional language, 5 denoting high politeness and professionalism).

2. Identify all instances where the response lacks politeness or professionalism.

3. Offer suggestions for maintaining a respectful and professional tone.\\

Relevance (How well does the candidate’s response directly address and align with the given interview question?):

1. Rate the candidate's relevance to the interview question on a scale of 0 to 5 (0 representing a lack of relevance, 5 indicating a highly relevant response).

2. Identify all off-topic or irrelevant elements in the response.

3. Offer suggestions for maintaining focus and relevance in interview responses.\\

Conciseness (How effectively does the candidate deliver information in a brief, yet informative manner, avoiding unnecessary verbosity?):

1. Rate the candidate's conciseness on a scale of 0 to 5 (0 being overly verbose, 5 being appropriately concise).

2. Identify all instances where the response is excessively wordy or lacks brevity.

3. Offer suggestions for delivering information in a more concise manner.\\

 Grammaticality (How grammatically correct is the language used in the candidate’s response?):

1. Rate the candidate's grammatical accuracy on a scale of 0 to 5 (0 indicating numerous grammatical errors, 5 indicating impeccable grammar).

2. Identify all grammatical errors in the response.

3. Offer suggestions for grammatical improvements.\\

[NOTE: Please provide the evaluation report in the same format as above.]

}

\end{tcolorbox}

    \caption{ Prompt to evaluate a response to \textit{Question Type 4} across six human evaluation criteria.}
    \label{fig14}
\end{figure*}


\begin{figure*}
    \centering
    \begin{tcolorbox}[enhanced,fit to height=18cm,
  colback=blue!20!white!10!white,colframe= blue!60!white,title= GPT-3.5 Turbo evaluation of the response to \textit{Question Type 4}, drop fuzzy shadow]

  {\fontfamily{qcr}\selectfont
 
Assessment Criteria:\\

Fluency:

1. Rating: 4

2. Instances of disfluency: The response is generally fluent with minimal interruptions.

3. Improvement suggestions: Maintain the fluency and consider adding more natural transitions between ideas for smoother delivery.\\

Coherence:

1. Rating: 4

2. Instances where the response lacks coherence: The response is coherent, but there's a slight lack of detail.

3. Improvement suggestions: Expand on your points to provide more depth and context. Ensure a logical flow between ideas.\\

Tone/Politeness:

1. Rating: 4

2. Instances where the response lacks politeness or professionalism: The tone is polite and professional throughout.

3. Improvement suggestions: Maintain the respectful tone and consider expressing enthusiasm about engaging with the CEO.\\

Relevance:

1. Rating: 3

2. Instances of off-topic or irrelevant elements: The response is relevant but could benefit from more specifics.

3. Improvement suggestions: Provide concrete examples of how you see your role contributing to the CEO's vision. Add more depth to your thoughts on the company's future.\\

Conciseness:

1. Rating: 3

2. Instances of excessive verbosity: The response is concise but could be more succinct.

3. Improvement suggestions: Trim unnecessary words and aim for a more streamlined expression of ideas.\\

 Grammaticality:

1. Rating: 4

2. Instances of grammatical errors: The response is grammatically sound with only minor errors.

3. Improvement suggestions: Pay attention to articles and prepositions for further improvement.\\

}
  
\end{tcolorbox}

    \caption{GPT-3.5 Turbo evaluation of the response to \textit{Question Type 4}.}
    \label{fig15}
\end{figure*}


\begin{figure*}
    \centering
    \begin{tcolorbox}[enhanced,fit to height=22cm,
  colback=blue!20!white!10!white,colframe= blue!60!white,title= GPT-4 Turbo evaluation of the response to \textit{Question Type 4}, drop fuzzy shadow]

  {\fontfamily{qcr}\selectfont
 
Assessment Criteria:\\

Fluency:

1. Rating: 2

2. Instances of disfluency: The candidate's response lacks smooth transitions, and there are instances of repetition, affecting overall fluency.

3. Improvement suggestions: To improve fluency, the candidate should focus on expressing ideas more coherently, avoiding unnecessary repetition, and practicing the response to eliminate pauses.\\

Coherence:

1. Rating: 2

2. Instances where the response lacks coherence: The response lacks a clear structure, making it challenging to follow the logical flow of ideas. Repetition of phrases also contributes to a lack of coherence.

3. Improvement suggestions: The candidate should organize their thoughts logically, perhaps by structuring the response into distinct sections such as discussing the CEO's vision, the candidate's role, and seeking feedback. Additionally, avoiding redundant phrases will enhance coherence.\\

Tone/Politeness:

1. Rating: 4

2. Instances where the response lacks politeness or professionalism: The tone is generally polite, but the response could benefit from more formal language to enhance professionalism.

3. Improvement suggestions: Consistently maintaining a formal tone and expressing gratitude for the opportunity to discuss these matters with the CEO would enhance the overall professionalism of the response.\\

Relevance:

1. Rating: 2

2. Instances of off-topic or irrelevant elements: The response is somewhat vague, lacking specific details about the candidate's vision or the company's future. The mention of seeking feedback seems premature and may not be the most relevant point to discuss initially.

3. Improvement suggestions: The candidate should provide more concrete examples and details, focusing on the company's current challenges and how their skills align with addressing those challenges.\\

Conciseness:

1. Rating: 2

2. Instances of excessive verbosity: The response is overly verbose, with unnecessary repetition, leading to a lack of conciseness.

3. Improvement suggestions: The candidate should aim for brevity by expressing ideas more directly. Eliminating redundant phrases and focusing on key points will contribute to a more concise and impactful response.\\

 Grammaticality:

1. Rating: 3

2. Instances of grammatical errors: While the candidate generally demonstrates acceptable grammatical accuracy, there are instances of awkward phrasing and minor grammatical errors.

3. Improvement suggestions: The candidate should review sentence structure to ensure ideas are expressed clearly. Proofreading can help catch and correct minor grammatical errors.\\

}
  
\end{tcolorbox}

    \caption{GPT-4 Turbo evaluation of the response to \textit{Question Type 4}.}
    \label{fig16}
\end{figure*}


\begin{figure*}
    \centering
    \begin{tcolorbox}[enhanced,fit to height=22cm,
  colback=green!20!white!10!white,colframe= blue!60!white,title= Human evaluation of the response to \textit{Question Type 4}, drop fuzzy shadow]

  {\fontfamily{qcr}\selectfont
 
Assessment Criteria:\\

Fluency:

1. Rating: 3

2. Instances of disfluency: The repetition of "about" and "I will discuss" disrupts the flow (e.g., "And I will discuss about my role in this vision"). The phrase "and I will tell him about the what I think about the future of the company" is awkward and disrupts fluency.: "

3. Improvement suggestions: Practice speaking in a structured manner, and using varied sentence constructions to avoid repetition. Prepare key points in advance to ensure a smoother delivery. Consider rephrasing it as "I would discuss his vision for the company's future and explore how my role aligns with this vision." \\

Coherence:

1. Rating: 3

2. Instances where the response lacks coherence: The response is somewhat repetitive and could be more logically structured. The phrase "And I will tell him about the what I think about the future of the company" is unclear and confusing.

3. Improvement suggestions: Organize thoughts before speaking. Start with the CEO's vision, then transition to discussing your role, followed by your own perspective. Ensure each point is clearly articulated and connected to the previous one.\\

Tone/Politeness:

1. Rating: 4

2. Instances where the response lacks politeness or professionalism: The response is generally polite but lacks formality in phrasing.

3. Improvement suggestions: Please be sure to use more formal language when discussing potential interactions with senior leadership. Consider rephrasing it to "I would be interested in hearing about his vision for the company's future and my role within that framework."\\

Relevance:

1. Rating: 4

2. Instances of off-topic or irrelevant elements: The response is relevant, but the mention of seeking feedback on personal performance may not be appropriate in a first meeting with a CEO.

3. Improvement suggestions: Focus on the CEO’s vision, the company’s future, how you can contribute to it, and discuss personal performance in a more appropriate setting.\\

Conciseness:

1. Rating: 3

2. Instances of excessive verbosity:  The response includes unnecessary repetition (e.g., "about the what I think about the future of the company").

3. Improvement suggestions: Eliminate repetitive phrases and focus on delivering each point succinctly. Consider rephrasing as "I would discuss the company's future vision and my role within that framework"\\

 Grammaticality:

1. Rating: 3

2. Instances of grammatical errors:  Incorrect use of "about" in "I would like to discuss about his vision" (should be "discuss his vision"). The phrase "And I will tell him about the what I think about the future of the company" is grammatically incorrect.

3. Improvement suggestions: Review grammar rules related to prepositions and sentence structure. Consider rephrasing the sentence to "I would discuss his vision for the company's future and my role in realizing that vision."\\

}
  
\end{tcolorbox}

    \caption{Human evaluation of the response to \textit{Question Type 4}.}
    \label{fig17}
\end{figure*}

\end{document}